\documentclass[a4paper,fleqn]{cas-dc}

\usepackage[numbers]{natbib}
\usepackage{subfig}
\usepackage{float}
\usepackage{array}   

\begin{document}
\def\floatpagepagefraction{1}
\def\textpagefraction{.01}
\shorttitle{Tao Meng et~al. Knowledge-Based Systems}
\shortauthors{Wang et~al.}

\title [mode = title]{Dynamic Fusion-Aware Graph Convolutional Neural Network for Multimodal Emotion Recognition in Conversations}


\author[1]{Tao Meng}

\author[1]{Weilun Tang}

\author[1]{Yuntao Shou}

\author[1]{Yilong Tan}

\author[1]{Jun Zhou}

\author[1]{Wei Ai}
\cormark[1]

\author[2]{Keqin Li}

\cortext[1]{Corresponding author: aiwei@hnu.edu.cn}

\address[1]{College of Computer and Mathematics, Central South University of Forestry and Technology, 410004, Hunan, Changsha, China}

\address[2]{Department of Computer Science, State University of New York, New Paltz, New York, 12561, USA}

\begin{abstract}
   Multimodal emotion recognition in conversations (MERC) aims to identify and understand the emotions expressed by speakers during utterance interaction from multiple modalities (e.g., text, audio, images, etc.). Existing studies have shown that GCN can improve the performance of MERC by modeling dependencies between speakers. However, existing methods usually use fixed parameters to process multimodal features for different emotion types, ignoring the dynamics of fusion between different modalities, which forces the model to balance performance between multiple emotion categories, thus limiting the model's performance on some specific emotions. To this end, we propose a dynamic fusion-aware graph convolutional neural network (DF-GCN) for robust recognition of multimodal emotion features in conversations. Specifically, DF-GCN {integrates} ordinary differential equations into graph convolutional networks {(GCNs)} to {capture} the {dynamic} nature of emotional dependencies {within} utterance interaction networks and leverages the prompts generated by the global information vector (GIV) of the utterance to guide the dynamic fusion of multimodal features. This allows our model to dynamically change parameters when processing each utterance feature, so that different network parameters can be equipped for different emotion categories in the inference stage, thereby achieving more flexible emotion classification and enhancing the generalization ability of the model. {Comprehensive} experiments {conducted} on two public multimodal conversational datasets {confirm} that the proposed DF-GCN model {delivers} superior performance, {benefiting significantly from the dynamic fusion mechanism introduced.} To the best of our knowledge, this is the first framework that adaptively assigns different fusion weights to different emotion categories during inference, leading to more effective multimodal information integration. Our code is available at \href{https://github.com/yuntaoshou/DFGCN}{https://github.com/yuntaoshou/DFGCN}.
\end{abstract}

\begin{keywords}
\sep Multimodal Emotion Recognition
\sep Dynamic Fusion  
\sep Graph Convolutional Neural Network
\sep Prompt Learning
\end{keywords}

\maketitle

\section{INTRODUCTION}

Multimodal emotion recognition in conversations (MERC) aims to exploit multimodal information (e.g., text, audio, video, etc.) in the dialogues to understand and identify the emotional state of the speaker \cite{ma2023transformer, lu2025lecm, wei2025emotional, shou2026comprehensive, shou2022conversational, shou2024adversarial, shou2024low}. MERC has become an important research topic that has attracted much attention due to its wide application in opinion mining \cite{estrada2020opinion, fu2025sdr, meng2024deep, shou2025masked}, healthcare \cite{pujol2019emotion, meng2024multi, shou2026graph}, psychological counseling \cite{zhou2020design, shou2024efficient, shou2025spegcl}, and building empathetic dialogue systems. MERC research not only helps us to deeply understand the emotional changes of users in multi-round dialogues but also improves the intelligence of dialogue systems, making them more capable of emotional perception and response. Unlike traditional non-conversational or unimodal emotion recognition (ER) methods, MERC focuses more on the fusion of semantic features from conversational context and multiple modalities \cite{khare2024emotion, ai2026paradigm, shou2025contrastive}. The core challenge is how to effectively model and fuse the emotional dependencies between the context and multimodality in the conversation \cite{chudasama2022m2fnet, shou2025revisiting, shou2025dynamic}.  Specifically, MERC must fully capture the subtle changes in emotions in the conversation, understand the interaction patterns between multimodality, and reasonably fuse features to achieve more accurate emotion recognition \cite{liu2023multi, shou2025gsdnet, shou2023graphunet}.

\begin{figure}
	\centering
	\includegraphics[width=1\linewidth]{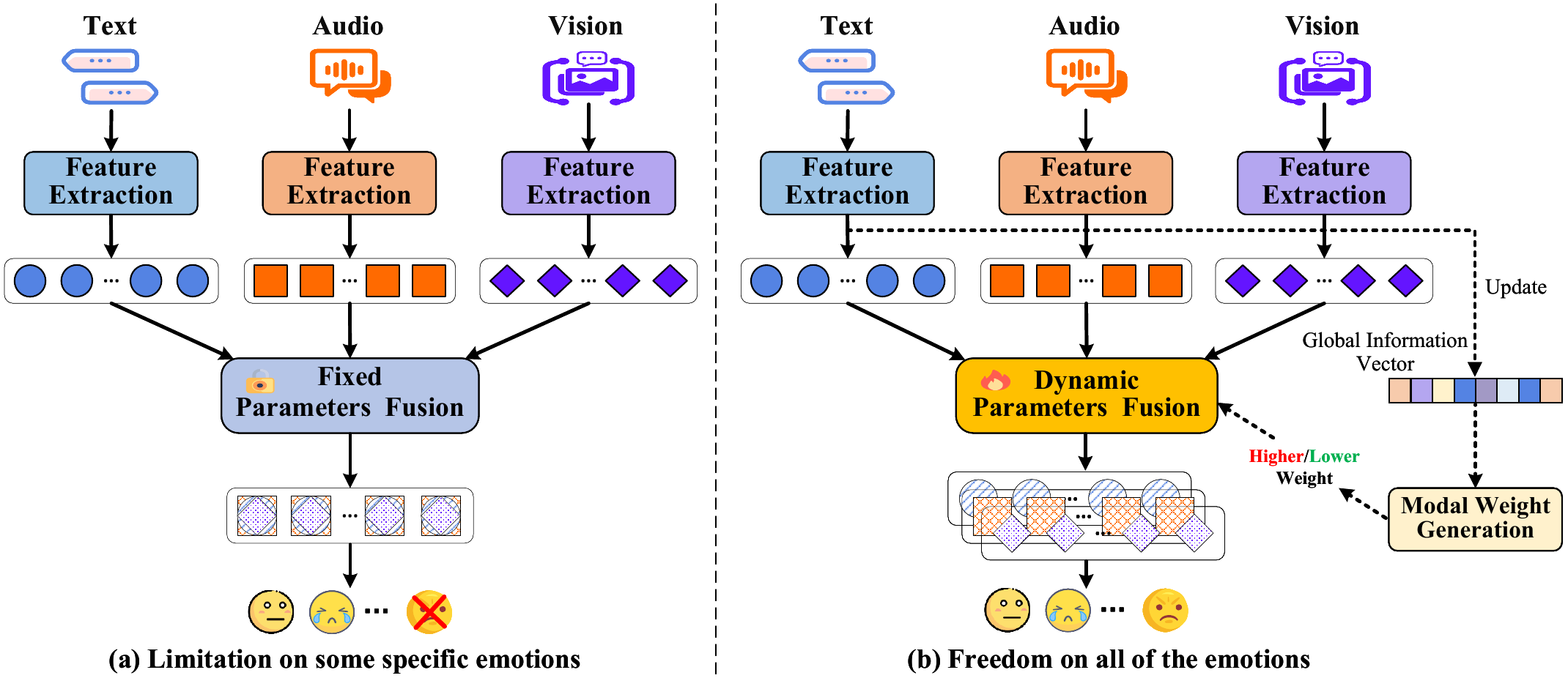}
	\caption{Illustration of the difference between traditional static fusion and dynamic fusion processes. (a) Previous studies usually use fixed network parameters to fuse multimodal features. (b) Our proposed method generates modal parameters through GIV, thereby achieving dynamic multimodal feature fusion.}
	\label{fig:figure1}
\end{figure}

The mainstream MERC methods mainly use a Transformer or GCN to model the emotional dependencies. and fuse the semantic features in the conversation \cite{tu2024adaptive,yang2024emotion,yi2024multimodal, shou2025graph}. For example, the Transformer-based self-distillation model (SDT) \cite{ma2023transformer} adaptively learns weights related to different modalities through self-distillation technology and a hierarchical gated fusion strategy to perform feature fusion better. TBJE \cite{delbrouck-etal-2020-transformer} leverages the Transformer architecture and a modular collaborative attention mechanism to effectively capture and integrate the key factors of emotion and sentiment expressed through multimodal language. DER-GCN \cite{ai2024gcn} constructs multiple emotional relationship graphs through events to capture the emotional dependencies between multimodalities in dialogue, and uses self-supervised learning and multi-information Transformer to optimize feature fusion. M3Net \cite{chen2023multivariate} leverages multi-frequency information within a graph to enhance the representation of multivariate relationships and capture complex relationships across modalities and context.

The above methods have achieved certain results in the MERC task through feature fusion. However, fixed parameters are often used in the inference stage to fuse the multimodal features across different emotion types. As shown in Figure \ref{fig:figure1}, although this unified parameter configuration simplifies the design of the model, the performance of the model on minority samples will drop significantly when faced with diverse emotion categories. In addition, due to the lack of specialized optimization for different emotion categories, the model has to balance performance between different categories, making it difficult to fully capture and represent the unique characteristics of each emotion. This not only affects the flexibility of the model but also limits its accuracy and sensitivity in certain emotion categories, making it difficult to achieve detailed distinction and recognition of complex emotions.

Therefore, MERC needs to consider the dynamics of semantic dependencies in the conversation context and adaptively equip different network parameters for the fusion of utterance features of different emotion categories as the starting point of model design. A key aspect of capturing this context is modeling the sequential relationships between utterances. While pre-trained language models like RoBERTa provide powerful utterance-level semantics, they lack built-in mechanisms for inter-utterance context. To this end, we incorporate a Bidirectional GRU (Bi-GRU) to efficiently capture these contextual dependencies in the conversation flow. Inspired by the above analysis, we propose a DF-GCN for robust recognition of multimodal emotion features in conversations. The core of DF-GCN is the design of two serial graph convolution blocks, namely, the Static Graph Convolution  (SGCODE) block and the Dynamic Graph Convolution (DGCODE) with ordinary differential equations block. Specifically, we first construct an emotional interaction graph based on multimodal utterance features and input it into SGCODE to capture the dynamic emotional dependencies between multimodal utterances. Then, the multimodal utterance features with emotional dependencies are input into the Transformer to obtain the GIV, and the multi-layer perceptron is used to generate prompts to obtain the dynamic weight of the utterance. Finally, the dynamic weight and multimodal utterance features are input into DGCODE for dynamic fusion to complete the emotional recognition of the utterance. It is worth noting that in the inference stage of the model, the parameters of the DGCODE we constructed are dynamically changed so that different network parameters can be equipped for utterances of different emotion categories in the inference stage, to achieve more flexible emotion classification and enhance the generalization ability of the model. The main contributions of this paper are summarized as follows:

\begin{itemize}
    \item We propose a DF-GCN for MERC. DF-GCN can adaptively equip utterances of different emotion categories with different network parameters to achieve more flexible emotion classification.

    \item We construct a novel prompt generation network to obtain GIV and dynamic fusion weights with rich semantics to guide the network to dynamically fuse utterance features with different emotion categories effectively.


    \item Extensive experiments were conducted on the IEMOCAP and MELD datasets, and the results verified the effectiveness of DF-GCN, especially in terms of WA and WF1, which significantly surpassed existing mainstream methods.
\end{itemize}

\section{RELATED WORK}

\subsection{Multimodal Emotion Recognition}
Multimodal emotion recognition in conversation (MERC) requires processing and analyzing speech, facial expressions, and text data to discern and understand human emotional states.
\cite{saxena2020emotion, shou2025cilf}.
In this paper, we roughly classify MERC methods into three categories: recurrent neural networks (RNNs), Transformers, and GCN.

RNN-based MERC methods mainly extract contextual semantic information from multimodal features through recurrent memory units. For example, DialogueRNN \cite{majumder2019dialoguernn} considers emotions in conversation by constructing GRUs. The CNN-RNN-based methods \cite{kollias2020exploiting} combine multiple CNN features with RNN subnetworks through pre-training and feature fusion to achieve dimensional emotion recognition.

Transformer-based MERC methods effectively extract and integrate emotional features from different modalities through self-attention mechanisms and multimodal fusion strategies.
For example, the Transformer-based self-distillation model (SDT) \cite{ma2023transformer} aims to apply self-distillation technology and hierarchical gated fusion strategy to adaptively learn weights related to different modalities to effectively capture intra-modal and inter-modal interactions in dialogue data. The main modality transformer (MMTr) \cite{zou2022improving} learns information interaction between modalities through the multi-head attention mechanism to enhance the representation of weak modalities. {HTNet \cite{wang2025visual} improves the accuracy of micro-expression recognition by fusing CNN and Transformer to achieve joint spatial-temporal modeling of micro-expressions.}

GCN-based MERC methods mainly model the interaction relationships among multiple modalities in the dialogue through graph structure and captures and analyzes the complex dependencies between these modalities through the graph convolutional network.
For example, Dialoguegcn \cite{ghosal2019dialoguegcn} captures the contextual information of the dialogue by modeling the mutual dependencies between each discourse in the dialogue.
C-GCN \cite{nie2020c} constructs association graphs and uses intra-class and inter-class correlations between videos for feature learning and information fusion.
DER-GCN \cite{ai2024gcn} captures the dependencies between speakers and events in a conversation by constructing multiple relationship graphs and uses self-supervised learning and multi-information transformers to optimize feature fusion and loss functions. By introducing context-aware graphs and knowledge graphs into graph convolutional networks \cite{fu2022context}, multimodal feature representations are enhanced, enabling more accurate emotion recognition.

Although these methods have achieved good performance on MERC, they often rely on fixed parameters to fuse multimodal features across different emotion types during the inference stage. Specifically, due to the limitation of fixed parameters, it is difficult for the model to adaptively adjust according to the specific characteristics of each emotion type, resulting in the inability to fully explore and express the unique characteristics of each emotion. This limitation prevents models from accurately capturing the subtle differences and nuances among emotions when dealing with complex emotional situations, thus affecting the accuracy of emotion recognition and the generalization ability of the model.

\begin{figure*}
	\centering
	\includegraphics[width=1\linewidth]{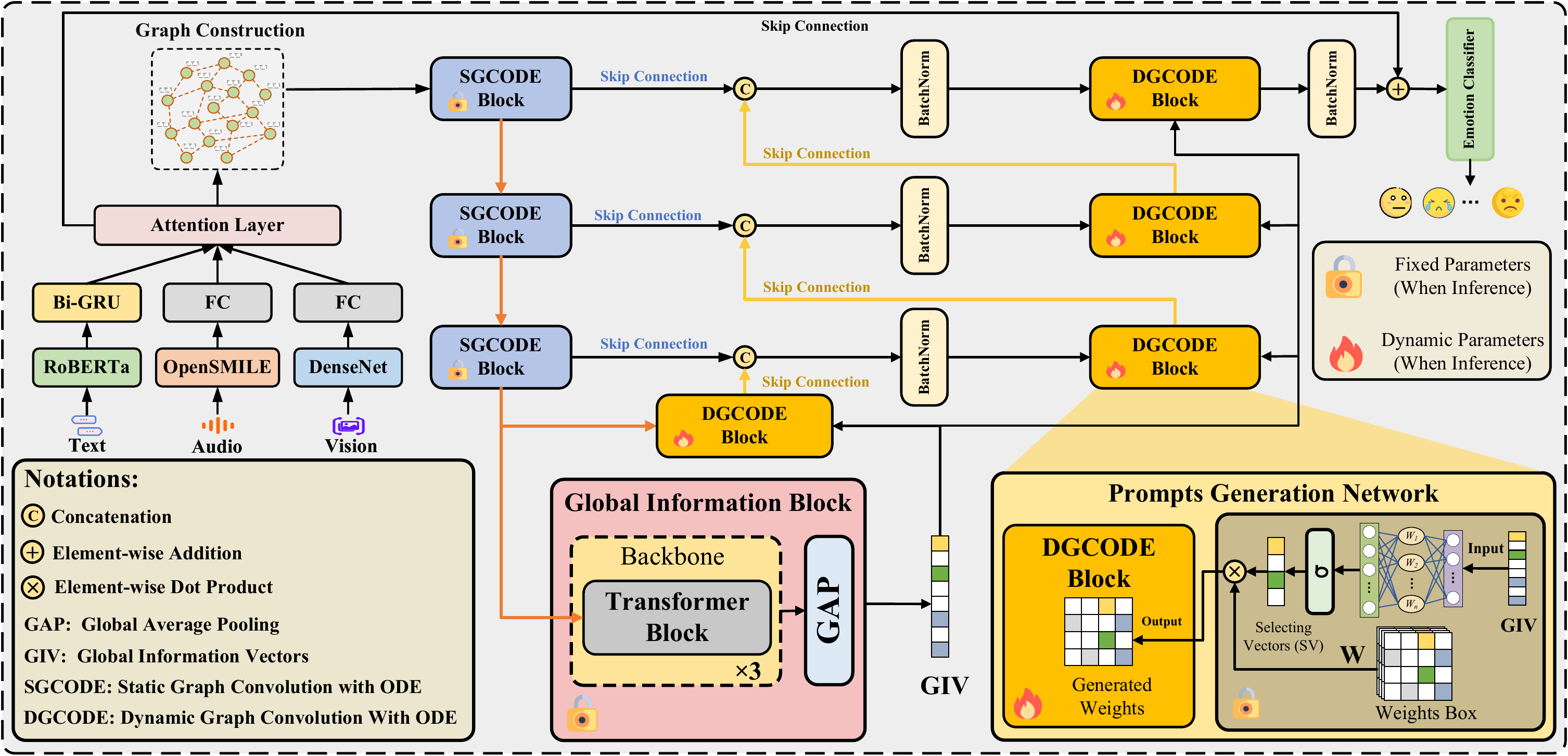}
	\caption{The overall architecture of the proposed DF-GCN model. The framework consists of three main components: (i) Graph Construction, where multimodal utterances from text (RoBERTa), audio (OpenSMILE), and vision (DenseNet) are encoded and integrated through an attention layer and Bi-GRU to form discourse-level interaction graphs; (ii) Global Information Block, which employs stacked Transformer layers and global average pooling to generate GIV that guide context-aware fusion; and (iii) Prompts Generation Network, where DGCODE blocks dynamically generate weight prompts for adaptive multimodal feature integration. SGCODEcaptures structural dependencies, while DGCODE models temporal and contextual dynamics. Skip connections and batch normalization enhance training stability, and the final fused representations are used for emotion classification.}
	\label{fig:figure2}
\end{figure*}

\subsection{Prompt Learning}
In natural language processing (NLP), prompt-based techniques are effective ways to provide contextual information to models for fine-tuning on target tasks \cite{niu2024lightzero}. This approach provides task-specific information to models through prompts, enabling them to adjust their output behavior according to the context \cite{guo2024scattering}. However, unlike traditional manually designed instruction sets, learnable prompts can automatically adapt to task requirements through training and adjust model parameters more efficiently \cite{zhou2022learning}. Compared with manually written static instructions, learnable prompts not only have advantages in adaptability but also can more fully exploit the potential of the model performance and computational cost. As prompt learning techniques can simulate task-specific contextual information, they have not only made significant progress in NLP, but have also been successfully applied to other fields (e.g., fine-tuning to visual tasks \cite{li2021prefix,jia2022visual,khattak2023maple}, and incremental learning \cite{smith2023coda,wang2022dualprompt}. In these applications, prompts act as a bridge between pre-trained models and downstream tasks, enabling models to respond more flexibly to changing task requirements. In addition, prompt learning techniques have also shown great potential in multi-task learning scenarios \cite{he2022hyperprompt}. The correct choice of prompts for different tasks is crucial because the contextual information required for each task may be different, which has a direct impact on the performance of the model. For example, the LLM (large language model)-based MERC method,such as leveraging visual and textual prompts to guide emotion understanding within a large visual language model \cite{wang2024htnet}, improves multimodal emotion recognition performance. Visual Prompting enhances cross-modal emotion recognition capabilities within large language models \cite{zhang2024visual} and improves recognition performance in complex scenarios.

\subsection{Continuous Graph Neural Networks}

Neural Ordinary Differential Equations (ODEs) represent a recent advance in modeling continuous-time dynamic systems \cite{chen2018neural}. Instead of relying on the discrete stacking of hidden layers as in conventional neural networks, neural ODEs parameterize the derivatives of hidden states with a neural function, enabling continuous inference along the temporal dimension. This paradigm provides a more precise characterization of temporal evolution and is particularly well-suited for tasks that involve complex time-varying behaviors. Building upon this idea, Continuous Graph Neural Networks (CGNNs) \cite{xhonneux2020continuous} extend the ODE framework to graph-structured data. CGNNs introduce a continuous message-passing mechanism, where the evolution of node representations is governed by an ordinary differential equation rather than a fixed number of propagation layers. This allows information to flow across nodes in a continuous manner, alleviating the constraints imposed by discrete architectures. To further address the problem of over-smoothing commonly observed in graph neural networks, CGNN incorporates a restart distribution mechanism, which periodically reset node representations to initial states during propagation. This design preserves informative distinctions among nodes while maintaining the benefits of continuous-time modeling.

\section{Preliminaries}

\subsection{Graph Neural Networks}

{Consider} a graph $G = (V, E)$, where $V$ {denotes the} set of nodes and $E$ represents {the} set of edges. {The goal is to derive meaningful representations for each node within the graph.} Nodes in $V$ are represented by a set of features stored in the node feature matrix $X \in \mathbb{R}^{|V|\times d}$, where $d$ represents the dimension of the feature vector. Each row of $X$ corresponds to the feature representation of node $v \in V$. To model the connectivity between nodes in the graph, we utilize the binary adjacency matrix $A \in \mathbb{R}^{|V|\times |V|}$, where each element $a_{ij}$ represents whether there is an edge between node $i$ and node $j$. Specifically, $a_{ij} = 1$ indicates that there is an edge between nodes $i$ and $j$, while $a_{ij}$ = 0 indicates that there is no edge between the two nodes. The adjacency matrix $A$ represents the dependency relationship between nodes. The main goal of the graph learning task is to learn a node representation matrix $H$ of size $|V|\times k$, where $k$ is the dimension of the learned node embedding.

\subsection{Neural Ordinary Differential Equation}

{ODEs offer an innovative approach to continuous-time dynamic modeling, where the forward propagation process in neural networks is framed as a solution to differential equations.} Unlike traditional neural network architectures, neural differential equations can simulate more complex and continuous-time dynamic behaviors by directly describing the evolution of input data as a dynamic system of differential equations. Specifically, consider an input data $x(t)$, whose differential equation evolving over time $t$ is defined as follows:
\begin{equation}
    \frac{d}{dt}x(t) = f(x(t), t, \theta)
\end{equation}
where $x(t)$ is the state at time $t$, and $f(x(t), t, \theta)$ is a function that describes the evolution of the system, usually approximated by a neural network.

\section{Methodology} 
The proposed DF-GCN contains five modules: multimodal feature encoding, static graph convolution, global information vector generation, dynamic graph convolution, and emotion classification. The detailed architecture of the proposed DF-GCN is shown in Fig. \ref{fig:figure2}.

\subsection{Task Definition}
In MERC, the number of speakers $M$ ($M$ $\geq$ 2) is not fixed. We assume that $M$ speakers participate in the conversation, represented as $S = \{S_1, S_2, ..., S_M \}$. The utterances of the speakers in the conversation are ordered in time, represented as $T= \{t_1, t_2, ..., t_N\}$, where $N$ represents the total number of utterances. The task of MERC is to identify the emotion label (e.g., happy, sad, frustrated, neutral, angry, and depressed) of each utterance in $T$.
   
\subsection{Multimodal Feature Encoding}
Following the previous work \cite{ma2023transformer}, we first use the RoBERTa \cite{kim2021emoberta} Large model, the openSMILE \cite{eyben2013recent} toolkit, and the DenseNet \cite{huang2017densely} model to encode the text, audio, and video data in each utterance respectively, and obtain the initial embedding representation as ${X_t}$, ${X_a}$, and ${X_v}$.

Then, we further use Bi-GRU to capture the contextual information of text features and use fully connected (FC) networks to transform audio and video features, respectively, to facilitate the understanding of spatiotemporal information. The formula is as follows:
\begin{equation}
\begin{aligned}
	\mathbf{h}_{t}=\overleftrightarrow{\mathrm{Bi\mbox{-}GRU}(X_{t})}, \mathbf{h}_{a}=\overleftrightarrow{\mathrm{FC}(X_{a})},
    \mathbf{h}_{v}=\overleftrightarrow{\mathrm{FC}(X_{v})}.
    \end{aligned}	
\end{equation}

To obtain preliminary fused multimodal features, we define a learned attention weight $\alpha_t$, $\alpha_a$, $\alpha_v$ for each modality to represent their importance. For each modality, a linear transformation is used to compute its attention score. This is done by computing the dot product of $\mathbf{h}_{m}$ as follows:
\begin{equation}
\begin{aligned}
	\mathbf{\alpha}_{m}  &=\mathbf{q}^{T}\mathbf{h}_{m}, \\
	\hat{\alpha}_{{m}}  &={\frac{\exp(\mathrm{\alpha}_{m})} {\exp(\mathrm{\alpha}_{t})+\exp(\mathrm{\alpha}_{a})+\exp(\mathrm{\alpha}_{v})}}
     \end{aligned}
\end{equation}
where $m \in \{ t, a, v\}$ represents any modality from text, audio, or vision, $\mathbf{q}$ is a learnable query vector used to compute the attention score for each modality, with the same dimension as the modality feature $\mathbf{h}_m$, $\hat{\alpha}_{{m}}$ represents the attention score after normalization of the $m$-th modality. The scores are converted into attention weights using the softmax function to ensure that the sum of the weights is 1. Then, the features from different modalities are weighted and summed to obtain the fused multimodal feature representation $\mathbf{h}_{f}$.

To align heterogeneous multimodal features within a unified semantic space, we adopt an attention-based soft alignment strategy. Instead of requiring strict temporal or frame-level correspondence, the learnable query vector dynamically evaluates the contribution of each modality to the emotional semantics of the current utterance. Modalities that provide more emotionally salient cues receive higher attention weights and therefore dominate the fused representation. This semantic-level alignment allows the model to robustly integrate multimodal information while mitigating noise or inconsistencies across modalities. The fused feature representation is then obtained through a weighted sum as follows:
\begin{equation}
\begin{aligned}  \mathbf{h}_{f}=\hat{\alpha}_{t}\mathbf{h}_{t}+\hat{\alpha}_{\mathrm{a}}\mathbf{h}_{a}+\hat{\alpha}_{v}\mathbf{h}_{v}.
\end{aligned}
\end{equation}

\subsection{Static Graph Convolution}

Traditional Graph Convolutional Networks (GCNs) {aggregate} information in a discrete, {layer-wise fashion, updating node representations over a fixed number of propagation steps.} Such a formulation implicitly assumes that information propagation occurs at discrete time points. However, in conversational emotion recognition, emotional states evolve continuously over time rather than through abrupt layer-wise transitions. Modeling emotional dependency propagation using discrete graph layers may therefore limit the ability to capture long-term and smooth temporal dynamics.

To address this issue, we aim to reinterpret the discrete information propagation process of GCNs as a continuous-time dynamical system. By doing so, emotional state evolution can be explicitly modeled as a temporal process governed by an ordinary differential equation (ODE), enabling more flexible and stable information propagation over extended time horizons.

Let $E$ denote the initial node embeddings produced by the encoder, $A$ the normalized adjacency matrix, and $W$ the learnable weight matrix. After multiple propagation steps, the accumulated node representation can be written as:
\begin{equation}
H_n = \sum_{n=1}^{N} \sum_{k=0}^{n} A^k E W^k.
\label{eq:discrete_gcn}
\end{equation}

To bridge discrete propagation and continuous dynamics, we observe that the summation over propagation depth can be interpreted as a Riemann sum over a temporal interval. Letting the propagation index approach a continuous variable $t$, Eq.~\eqref{eq:discrete_gcn} can be rewritten in integral form as:
\begin{equation}
H(t) = \frac{1}{N} \sum_{n=1}^{N} \int_{0}^{t+1} A^{s} E W^{s} \, ds. 
\label{eq:riemann_integral}
\end{equation}

This transformation ensures that the continuous-time formulation remains theoretically consistent with standard GCN behavior.

Taking the derivative of $H(t)$ with respect to time yields:
\begin{equation}
\frac{dH(t)}{dt} = \frac{1}{N} \sum_{n=1}^{N} A^{t+1} E W^{t+1}.
\label{eq:first_derivative}
\end{equation}

While this expression captures the instantaneous rate of information propagation, shallow aggregation may lead to information loss. To obtain a more stable formulation, we further analyze the second-order derivative:
\begin{equation}
\frac{d^2H(t)}{dt^2}
= \frac{1}{N} \sum_{n=1}^{N}
\left(
A^{t+1} E W^{t+1}
+ A^{t+1} E W^{t+1} \ln W
\right).
\label{eq:second_derivative}
\end{equation}

Rearranging and integrating both sides with respect to time leads to the following first-order ODE:
\begin{equation}
\frac{dH(t)}{dt} = \ln A \, H(t) + H(t) \ln W + c. 
\label{eq:ode_general}
\end{equation}
where $c$ is a constant determined by the initial condition.

Let the eigen-decompositions of the adjacency matrix and weight matrix be given by:
\begin{equation}
A = P \Lambda P^{-1}, \quad
W = Q \Phi Q^{-1}.
\end{equation}

The initial condition $H(0)$ is defined as:
\begin{equation}
(P^{-1} H(0) Q)_{ij} =
\frac{\bar{A}_{ii} \bar{E}_{ij} \bar{\Phi}_{jj} - \bar{E}_{ij}}
{\ln \bar{A}_{ii} + \ln \bar{\Phi}_{jj}}. 
\label{eq:initial_condition}
\end{equation}

When $t = 0$, we obtain:
\begin{equation}
\left. \frac{dH(t)}{dt} \right|_{t=0} = A E W. 
\label{eq:initial_derivative}
\end{equation}

Combining Eqs.~\eqref{eq:ode_general}--\eqref{eq:initial_derivative}, the constant term can be derived as:
\begin{equation}
c = P \bar{E} Q^{-1} = E. 
\label{eq:constant_term}
\end{equation}

Substituting Eq.~\eqref{eq:constant_term} into Eq.~\eqref{eq:ode_general}, the final Graph ODE is obtained:
\begin{equation}
\frac{dH(t)}{dt}
= \frac{1}{N} \sum_{n=1}^{N}
\left(
\ln \hat{A} \, H(t) + H(t) \ln W + E
\right).
\label{eq:final_ode}
\end{equation}

The Graph ODE in Eq.~\eqref{eq:final_ode} describes the continuous evolution of node representations over time. In practice, this ODE is solved numerically using standard ODE solvers (e.g., the Runge--Kutta method):
\begin{equation}
H(t) = \mathrm{ODESolver}
\left(
\frac{dH(t)}{dt}, \, H_0, \, t
\right).
\label{eq:ode_solver}
\end{equation}

This continuous-time formulation enables smooth and long-range emotional dependency propagation and serves as the theoretical foundation for the subsequent dynamic graph convolution module.

The Graph ODE derived above provides a principled continuous-time formulation of graph information propagation that is theoretically consistent with discrete GCNs. To instantiate this formulation within our model, we parameterize the vector field of the ODE using a graph convolution operator, resulting in the proposed SGCODE. Specifically, we adopt the normalized adjacency matrix to encode structural dependencies among utterances and employ a learnable weight matrix that remains fixed during inference to ensure stable and consistent message propagation. Under this setting, the continuous evolution of node representations is governed by a static ODE system, where the graph topology and convolutional parameters jointly determine the propagation dynamics. By grounding SGCODE in the derived Temporal Graph ODE, we ensure that static graph convolution is not merely an architectural choice but a direct realization of the continuous-time dynamics implied by discrete GCN propagation. This static ODE-based formulation serves as the foundational encoding stage of our framework, upon which more flexible and context-adaptive dynamic graph convolution mechanisms are subsequently built.


\textbf{Emotional Interaction Graph Construction.} 
For node construction, we take the multi-modal data as the input. We represent each utterance $u_i$ as a node
$v_i$. For edge construction, we consider each node $v_i$ only interacts with the nodes within the context window $\{v_j\}_{j \in [\max(i-w, 1), \min(i+w, L)]}$, where $w$ is set to 10, $L$ is the number of utterances. We use $e_{ij}$ to represent the
edge from $v_i$ to $v_j$. For edge type, we assign each edge $e_{ij}$ with a speaker type identifier $\alpha_{ij}(p_{s(u_i)} \rightarrow p_{s(u_j)})$, where $p_{s(u_i)}$ and $p_{s(u_j)}$ represent the speaker identity of $u_i$ and $u_j$,
respectively.

The construction of the graph mainly depends on the adjacency matrix $\mathbf{A}$. The element \(\mathbf{A}_{ij}\) of the adjacency matrix represents the intensity of emotional dependence between multimodal utterances $i$ and $j$. Assuming that the features of utterances $i$ and $j$ are \(\mathbf{h}_i\) and \(\mathbf{h}_j\), the adjacency matrix $\mathbf{A}$ is constructed by calculating the cosine similarity between them. The formula is as follows:
\begin{equation}
\begin{aligned}
    \operatorname{sim}(\mathbf{h}_{i},\mathbf{h}_{j})={\frac{\mathbf{h}_{i}^{\top}\mathbf{h}_{j}}{\|\mathbf{h}_{i}\|\|\mathbf{h}_{j}\|}},{\mathbf{A}}_{i j}=\mathbb{I}(\operatorname{sim}(\mathbf{h}_{i},\mathbf{h}_{j})>\theta),
\end{aligned}
\end{equation}
where \(\mathbb{I}(\cdot)\) is the indicator function and \(\theta\) is the similarity threshold.

\textbf{Graph ODE Modeling.} Graph ordinary differential equation (ODE) is used to describe the changes in node features over time. The features of each node at time $t$ are updated through the graph ODE. In general, the form of graph ODE is as follows:
\begin{equation}
	{\frac{d\mathbf{h}_{i}(t+1)}{d t}}=f\left(\mathbf{h}_{i}(t),\mathbf{A},t\right),
\end{equation}
where \(\mathbf{h}_i(t)\) is the feature vector of node $i$ at time $t$. \(\mathbf{A}\) is the adjacency matrix of the graph. \(f(\cdot)\) is a GCN used to learn the relationship between nodes. Time $t$ represents the node features that change over time in a dynamic system.

\textbf{Convolution with Graph ODE.} Under the framework of graph ODE, the features of each node \(\mathbf{h}_i(t)\) are propagated through graph convolution according to the features of neighboring nodes and updated over time. The graph ODE models the changes in the features of each node in the form of time derivatives, while graph convolution provides information transfer between nodes. The formula for SGCODE is as follows:
\begin{equation}
	{\frac{d\mathbf{h}_{i}(t+1)}{d t}}=\sum_{j\in{\mathcal{N}}(i)}{\rm{ln}}\mathbf{A}_{i j}\mathbf{h}_{j}(t)+\mathbf{h}_{i}(t){\rm{ln}}W_{s},
\end{equation}
where $\mathcal{N}(i)$ represents the first-order neighbors of node $i$, and $W_{s}$ is a trainable parameter matrix that will remain fixed during the inference phase, $\sum_{j\in{\mathcal{N}}(i)}$ represents the summation of all nodes $j$ belonging to $\mathcal{N}(i)$.

\subsection{ Global Information Vector}
{Drawing inspiration from the Transformer’s ability to model global dependencies, we generate a GIV using global attention mechanisms and global average pooling (GAP), which guides the subsequent prompt generation and dynamic fusion process.}

\textbf{Global Context Modeling.} SGCODE obtains the representation of each utterance at time $t$ by simulating the evolution of the utterance representation of the graph over time \(\mathbf{h}_i(t)\). These utterance representations can be used as input for global attention to fuse important emotional information in the context of global conversation. Next, we calculate the attention weights and fuse the conversation context information:
\begin{equation}
\begin{aligned}
    \mathbf{A}_{score}&=\mathrm{softmax}\left({\frac{\mathbf{Q}\mathbf{K}^{T}} {\sqrt{d_{k}}}}\right), \\ \mathbf{Z}&=\mathbf{A}_{score}\,\mathbf{V},
\end{aligned}
\end{equation}
where $\mathbf{A}_{score}$ is the attention score matrix, which represents the relationship between each utterance and other utterances; \(d_k\) is the dimension of the query vector, which is used for scaling calculations; $\mathbf{Q}$, $\mathbf{K}$, and $\mathbf{V}$ denote the query, key, and value matrices obtained by linearly projecting the input features; and $\mathbf{Z}$ is the fused feature matrix of all utterances.

\textbf{Global Average Pooling.} The global information vector represents the global conversation information by averaging the features of all input utterances. The generated vector $\mathbf{g}$ can represent the global features of the entire emotional interaction graph. The GAP operation synthesizes the representation of all utterances into a fixed-length vector. The specific formula is as follows:
\begin{equation}
	\mathbf{g}=\frac{1}{N}\sum_{i=1}^{N}{\mathbf{z}}_{i},
\end{equation}
where $\mathbf{g}$ is the global information vector, representing the global features of the entire graph, and $\mathbf{z}_{i} \in \mathbf{Z}$ is the representation of each utterance output by the Transformer, $N$ represents the total number of utterances.

\subsection{Dynamic Graph Convolution}
In DGCODE, the core idea is to dynamically generate model parameters conditioned on the GIV. We clarify that the GIV, generated from the internal features of the input conversation, functions as an internal prompt or a conditioning signal. Unlike traditional prompt learning, which often employs external, human-designed task instructions, our internal prompt (GIV) encapsulates the global conversational context. This prompt then guides the Prompt Generation Network (PGN) to produce dynamic weights, effectively instructing the model on how to adaptively fuse multimodal features for the current emotional and contextual scenario. This conditioned dynamic fusion is applied to the graph ODE. Consequently, the ODE system in DGCODE is not a fixed dynamic system but one whose evolution is explicitly instructed by the unique context of the input, allowing it to equip different network parameters for different emotion categories during inference.

\textbf{Dynamic Weight Generation.} The PGN generates parameters related to the dynamic evolution of graph ODE based on the GIV, including node update rules and edge influences. Specifically, we first input the global information vector $\mathbf{g}$ into the MLP with a fully connected layer and use the activation function to perform nonlinear changes to obtain the selection vector $\mathbf{s}$:
\begin{equation}
	\mathbf{s}=\sigma ({w_l}\mathbf{g} + {b_l}),
\end{equation}
where ${w_l}$, ${b_l}$ are the trainable parameters. Then, we introduce the learnable weights box $W_b$ and perform dot product with the selection vector $\mathbf{s}$ to generate a dynamic weight matrix, denoted as $W_d$:
\begin{equation}
    {W_d} = \mathbf{s} \otimes {W_b}.
\end{equation}

\textbf{Dynamic Convolution with Graph ODE.} Unlike SGCODE, which {employs} learnable weights {during} training and fixed weights during inference, DGCODE {utilizes dynamically generated weights for both training and inference phases, enabling the fusion and updating of utterance features.} In addition, to avoid the over-smoothing problem of GCN, we concatenate the outputs of each layer of SGCODE and DGCODE and batch normalize them to construct residual connections to improve the model's generalization ability. The formula for DGCODE is as follows:

\begin{equation}
\begin{aligned}
    &\frac{{d{{\bf{u}}_i}(t + 1)}}{{dt}} = \sum\limits_{j \in {\cal N}(i)} {\rm{l}} {\rm{n}}{{\bf{A}}_{ij}}{{\bf{u}}_j}(t) + {{\bf{u}}_i}(t){\rm{ln}}{W_d},\\
    &{\bf{u}}_i(t) = {\rm{BN}}\left( {{\rm{Concat}}\left( {{\bf{h}}_i(t-1),{\bf{u}}}_i(t-1) \right){\rm{; }}\gamma ,\beta } \right),
    \end{aligned}
\end{equation}
where ${W_d}$ is the generated dynamic parameter, ${{\bf{h}}_i}$ is the output feature of SGCODE, ${{\bf{u}}_i}$ is the output feature of DGCODE, $\rm{BN}(.)$ is the batch normalization operation, $\gamma$ is the learnable scaling parameter, and $\beta$ is the learnable offset parameter.

Unlike static graph convolution, where a fixed weight matrix is shared across all samples, $W_d$ dynamically modulates the node update process at each infinitesimal time step. Specifically, $W_d$ controls the relative importance of different feature dimensions during temporal propagation, thereby shaping the continuous-time evolution of node representations. As a result, the same graph structure can exhibit different propagation behaviors under different conversational contexts. The influence of $W_d$ on the final classification is accumulated throughout the ODE integration process. Since the node representations produced by DGCODE are directly fed into the emotion classifier, $W_d$ affects which semantic or modality-related features are amplified or suppressed before prediction, shaping the decision process rather than acting as a post-hoc transformation. Importantly, $W_d$ is not an unconstrained parameter. It is generated by the prompt generation network conditioned on the GIV, which summarizes the global emotional context of the conversation. This design allows $W_d$ to be interpreted as an emotion-aware convolutional kernel that adapts the ODE dynamics to different emotional trends. Moreover, while $W_d$ modulates node self-dynamics, the graph topology and neighborhood aggregation remain unchanged, preserving interpretability and ensuring that $W_d$ modifies how information is integrated over time rather than which information is propagated.

\subsection{Emotion Classifier} The feature ${{\bf{u}}_i}$ obtained by DGCODE is batch normalized and input into the linear layer through residual connection with the feature ${{\bf{h}}_i}$ output by the encoder, and then through the softmax layer to obtain the probability distribution $\mathbf{p}_i$ of the emotional label. The formula is defined as follows:
\begin{equation}
\begin{aligned}
    {{\bf{v}}_i} &= {\rm{Concat(}}{{\bf{h}}_i},{\rm{BN}}({{\bf{u}}_i} \rm{; } \gamma, \beta)),\\
    {{\bf{r}}_i} &= {\mathop{\rm ReLU}\nolimits} ({{\bf{v}}_i}{W_r } + {b_r }),\\
    {{\bf{p}}_i} &= {\mathop{\rm softmax}\nolimits} ({{\bf{r}}_i}{W_c} + {b_c}),
    \end{aligned}
\end{equation}
where ${W_r }$, ${b_r }$, ${W_c}$, ${b_c}$, $\gamma$ and $\beta$ are learnable parameters.

Finally, we use the argmax function to obtain the emotional label with the highest probability of utterance $i$ and use the classification cross-entropy loss for model training.
\begin{equation}
\begin{aligned}
    {{\hat y}_i} &= {\mathop{\rm argmax}\nolimits} ({{\bf{p}}_i}), \\
    \mathcal{L}_{ce} &= -\frac{1}{n}\sum_{i=1}^{n}y_{i}\log(\hat{y}_{i}),
    \end{aligned}
\end{equation}
where ${{y}_i}$ and ${{\hat y}_i}$ are the predicted and true emotion labels of utterance $i$, respectively.
\section{EXPERIMENTS}

\subsection{Database Used}
IEMOCAP \cite{busso2008iemocap} and MELD \cite{kamath2007model} are the two most commonly used databases in the MERC. They are widely used to evaluate the effectiveness of models.

The IEMOCAP dataset is collected by the SAIL laboratory of the University of Southern California. It is a multimodal conversation dataset that contains about 12 hours of multimodal data such as video, audio, facial motion capture, and transcribed text. It is recorded through the improvisation or scripted dialogues of actors. The data is annotated with emotion categories and dimension labels such as anger, happiness, and sadness, providing rich materials for multimodal emotion analysis.

The MELD dataset is derived from the EmotionLines dataset. It adds audio and visual modalities based on the plain text dialogues of ``Friends". It contains 1443 dialogues and 13708 sentences. Each sentence is annotated with one of the seven emotion labels,such as anger, disgust, sadness, positive, negative, and neutral emotion labels. It is used for emotion recognition in multi-party conversations. Both datasets have important value in multimodal emotion analysis.

\subsection{Evaluation Metrics}
In this section, we mainly explain the evaluation metrics to verify the effectiveness of the model in this paper. This paper mainly uses the following four evaluation metrics:
1) acuracy (ACC); 2) weighted accuracy (WA); 3) F1; 4) weighted F1 (WF1). To consider the problem of data set category imbalance, more comprehensively evaluate the model performance, and meet the actual application needs, we use WA and WF1 as the main evaluation metrics.

\subsection{Baselines}

To verify the superior performance of our proposed DFGCN method, we compare it with other comparative methods as follows:

\textbf{QMNN} \cite{li2021quantum} uniformly handles multimodal fusion and dialog context modeling through quantum analogy, and performs end-to-end training using complex-valued neural networks.

\textbf{A-DMN} \cite{xing2020adapted} reconstructs the emotion recognition in conversation task into a question-answering-like reasoning problem, and models the speaker's self-dependency (emotional inertia) and mutual dependency (emotional contagion) through a collaborative architecture.

\textbf{CoMPM} \cite{lee2021compm} transforms the Dynamic Memory Network (DMN) into a four-module architecture. It efficiently processes multimodal inputs using hierarchical convolutional fusion, models separately the global BiLSTM states and individual LSTM states, and directionally extracts emotional clues of target utterances through multi-round attention memory iteration.

\textbf{MGLRA} \cite{meng2024masked} iteratively refines multimodal features through a memory-enhanced recurrent alignment module, fuses semantics using a masked graph convolutional network, and finally classifies emotions through MLP.
 
\textbf{MMGCN} \cite{hu2021mmgcn} uniformly models the dependencies among text, audio, and visual modalities in conversations through a fully connected heterogeneous graph, enhances context modeling by combining speaker embeddings, and finally performs emotion fusion and classification via deep spectral domain GCN.
 
\textbf{AdaIGN} \cite{tu2024adaptive} models the temporal dependencies of multimodality in conversations through a directed heterogeneous graph, dynamically filters conflicting modal information by combining node/edge-level selection strategies, and outputs emotion classification by fusing dual-graph features through a graph-level selection strategy.

\textbf{DER-GCN} \cite{ai2024gcn} proposes a multi-relational emotional interaction graph network, which fuses dialogue relations and event relations. It jointly reconstructs node features and edge structures through a self-supervised masked graph autoencoder, dynamically aggregates cross-relational semantics using a multi-information Transformer, and alleviates the long-tail data problem with the assistance of contrastive learning loss optimization.

\textbf{RGAT} \cite{ishiwatari2020relation} proposes relational positional encoding, which integrates sequence positional information into the relational graph attention network for the first time, solving the problem that traditional graph models ignore temporal dependencies.

\textbf{CBERL} \cite{meng2024deep} generates minority samples that conform to the original distribution through a dual-path generative adversarial network, constructs a deep joint variational autoencoder, constrains the latent space via KL divergence, and fuses complementary semantics of multimodal modalities.

\textbf{M3Net} \cite{chen2023multivariate} constructs a hypergraph to model high-order multimodal relationships and designs low-pass/high-pass filters to separate emotional common/differential signals, aiming to address the issues of insufficient modeling of complex multivariate relationships and neglect of high-frequency emotional signals in multimodal conversation emotion recognition.
    
\textbf{CT-Net} \cite{lian2021ctnet} solves the three major bottlenecks in traditional methods (i.e., multimodal misalignment, lack of long-range dependencies, and neglect of speaker influence) through cross-modal temporal alignment, context attention mechanism, and speaker-aware fusion.

\textbf{EmoBERTa} \cite{kim2021emoberta} replaces the traditional complex multi-module architecture by adding speaker labels and separating historical/current/future utterances, that is, constructing input sequences that can perceive speaker identities and dialogue contexts.

\textbf{FrameERC} \cite{li2025frameerc} leverages graph framelet transforms and a dual-reminder fusion mechanism to capture both fine-grained emotional cues and balanced cross-modal interactions.

\textbf{HiMul-LGG} \cite{fu2025himul} proposes a hierarchical decision fusion-based local–global graph neural network for multimodal emotion recognition in conversation. It aligns multimodal features through a hierarchical fusion strategy, models speaker dependencies via a local–global graph structure, and enhances cross-modal interaction using a multi-head attention mechanism.

\begin{table*}[htbp]
\caption{The performance of different methods on IEMOCAP Dataset. The values reported in each cell represent the Acc/F1. The best result in each column is in bold.} 
\label{tab:iemocap}
\setlength{\tabcolsep}{6pt}{
\begin{tabular}{l|cc cc cc cc cc cc |cc}
\toprule[0.9mm]
\multirow{2}{*}{Method} & \multicolumn{2}{c}{Happy} & \multicolumn{2}{c}{Sad} & \multicolumn{2}{c}{Neutral} & \multicolumn{2}{c}{Angry} & \multicolumn{2}{c}{Excited} & \multicolumn{2}{c}{Frustrated} & \multicolumn{2}{c}{Average(w)} \\
\cline{2-15}
& Acc & F1 & Acc & F1 & Acc & F1 & Acc & F1 & Acc & F1 & Acc & F1 & Acc & F1  \\
\midrule[0.4mm]
A-DMN \cite{xing2020adapted} & 43.1 & 50.6 & 69.4 & 76.8 & 63.0 & 62.9 & 63.5 & 56.5 & \textbf{88.3} & 77.9 & 53.3 & 55.7 & 64.6 & 64.3 \\
RGAT \cite{ishiwatari2020relation} & 42.3 & 51.6 & 70.1 & 77.3 & 61.6 & 65.4 & 64.2 & 63.0 & 69.3 & 68.0 & 57.8 & 61.2 & 66.5 & 65.2 \\
QMNN \cite{li2021quantum} & 41.4 & 39.7 & 72.9 & 68.3 & 54.1 & 55.3 & 65.4 & 62.6 & 66.0 & 66.7 & 55.6 & 62.2 & 60.8 & 59.9 \\
CoMPM \cite{lee2021compm} & 55.4 & 60.7 & 78.3 & 82.2 & 64.5 & 63.0 & 66.2 & 59.9 & 77.4 & 78.2 & 57.3 & 59.5 & 64.2 & 67.3 \\
EmoBERTa \cite{kim2021emoberta} & 56.9 & 56.4 & 79.1 & \textbf{83.0} & 64.0 & 61.5 & 70.6 & 69.6 & 86.0 & 78.0 & 63.8 & 68.5 & 67.3 & 67.3 \\
CTNet \cite{lian2021ctnet} & 47.9 & 51.3 & 78.0 & 79.9 & 69.0 & 65.8 & 65.7 & 67.2 & 85.3 & \textbf{78.7} & 52.2 & 58.8 & 68.0 & 67.5 \\
MMGCN \cite{hu2021mmgcn} & 48.9 & 47.1 & \textbf{82.3} & 81.9 & 64.1 & 66.4 & 64.2 & 63.5 & 73.9 & 76.2 & 60.0 & 59.1 & 66.8 & 66.8 \\
M3Net \cite{chen2023multivariate} & 59.1 & 60.9 & 74.5 & 78.8 & 65.4 & 70.1 & 69.2 & 68.1 & 73.4 & 77.1 & 72.8 & 67.0 & 72.5 & 71.1 \\
MGLRA \cite{meng2024masked} & \textbf{62.9} & \textbf{63.5} & 81.1 & 81.5 & 70.9 & 71.5 & 60.2 & 61.1 & 74.4 & 76.3 & 69.2 & 67.8 & 71.3 & 70.1 \\
CBERL \cite{meng2024deep} & 58.8 & 67.3 & 63.3 & 72.8 & 56.4 & 60.8 & 75.3 & 73.5 & 70.3 & 70.8 & \textbf{78.2} & 71.2 & 69.4 & 69.3 \\
AdaIGN \cite{tu2024adaptive} & 51.1 & 53.0 & 80.5 & 81.5 & 72.3 & 71.3 & 63.3 & 65.9 & 72.8 & 76.3 & 63.5 & 67.8 & 72.1 & 70.7 \\
DER-GCN \cite{ai2024gcn} & 60.7 & 58.8 & 75.9 & 79.8 & 66.5 & 61.5 & 71.3 & 72.1 & 71.1 & 73.3 & 66.1 & 67.8 & 69.7 & 69.4 \\ 
FrameERC \cite{li2025frameerc} & 60.3  & 56.2 & 80.1 &  80.9 & 66.4  & 67.7 & \textbf{75.5}  & \textbf{78.6} & 70.0 &  69.7 & 71.3  & 70.8 & 70.0  &  70.7 \\
HiMul-LGG \cite{fu2025himul} & 53.2 & 54.0   &  79.4 & 79.9 & 71.2 & 71.7 &  69.3 & 67.6   & 71.1 & 72.0 &  69.1 &
68.5 & 70.1  &  70.2  \\
\midrule
DF-GCN & \textbf{62.9} & 62.0 & 72.6 & 81.3 & \textbf{72.9} & \textbf{71.9} & 69.1 & 70.7 & 86.6 & 75.5 & 70.1 & \textbf{68.6} & \textbf{73.4} & \textbf{72.2} \\
\bottomrule[0.7mm]
\end{tabular}}
\end{table*}

\begin{table*}[htbp]
\caption{The performance of different methods on MELD Dataset. The values reported in each cell represent the Acc/F1. The best result in each column is in bold.}
\label{tab:meld}
\setlength{\tabcolsep}{4.5pt}{
\begin{tabular}{l| cc cc cc cc cc cc cc |cc}
\toprule[0.9mm]
\multirow{2}{*}{Method} & \multicolumn{2}{c}{Neutral} & \multicolumn{2}{c}{Surprise} & \multicolumn{2}{c}{Fear} & \multicolumn{2}{c}{Sadness} & \multicolumn{2}{c}{Joy} & \multicolumn{2}{c}{Disgust} & \multicolumn{2}{c}{Anger} & \multicolumn{2}{c}{Average(w)} \\ \cline{2-17}
& Acc & F1 & Acc & F1 & Acc & F1 & Acc & F1 & Acc & F1 & Acc & F1 & Acc & F1 & Acc & F1  \\
\midrule[0.4mm]
A-DMN \cite{xing2020adapted} & 78.8 & 78.9 & 55.4 & 55.3 & 8.7 & 6.8 & 24.7 & 23.9 & 24.6 & 22.7 & 3.5 & 3.2 & 41.3 & 40.9 & 55.5 & 55.4 \\
RGAT \cite{ishiwatari2020relation} & 77.6 & 78.1 & 43.2 & 41.5 & 5.6 & 2.4 & 29.8 & 30.7 & 59.1 & 58.6 & 3.4 & 2.2 & 45.2 & 44.6 & 60.2 & 59.5 \\
QMNN \cite{li2021quantum} & 71.2 & 77.0 & 45.8 & 49.8 & - & - & 24.3 & 16.5 & 53.5 & 52.1 & - & - & 42.9 & 43.2 & 60.8 & 58.0 \\
CoMPM \cite{lee2021compm} & 79.1 & 82.0 & 51.5 & 49.2 & 5.2 & 2.9 & 33.6 & 32.3 & 62.3 & 61.5 & 8.4 & 2.8 & 42.7 & 45.8 & 64.3 & 63.0 \\
EmoBERTa \cite{kim2021emoberta} & 78.9 & \textbf{82.5} & 50.2 & 50.2 & 1.8 & 1.9 & 33.3 & 31.2 & \textbf{72.1} & 61.7 & 9.1 & 2.5 & 43.3 & 46.4 & 64.1 & 63.3 \\
CTNet \cite{lian2021ctnet} & 75.6 & 77.4 & 51.3 & 50.3 & 5.1 & 10.0 & 30.9 & 32.5 & 54.3 & 56.0 & 11.6 & 11.2 & 42.5 & 44.6 & 61.9 & 60.2 \\
MMGCN \cite{hu2021mmgcn} & 75.8 & 77.0 & 52.5 & 49.6 & 4.4 & 3.6 & 18.8 & 20.4 & 54.9 & 53.8 & 10.2 & 2.8 & 44.1 & 45.2 & 58.3 & 58.4 \\
M3Net \cite{chen2023multivariate} & 73.6 & 79.1 & 57.6 & 59.5 & 23.1 & 13.3 & 45.6 & 42.9 & 63.2 & 65.1 & 24.3 & 21.7 & 50.4 & 53.5 & 63.9 & 65.8 \\
MGLRA \cite{meng2024masked} & 78.2 & 80.8 & 59.8 & 59.5 & 0.0 & 0.0 & 30.8 & 27.8 & 68.5 & 66.5 & 0.0 & 0.0 & 43.8 & 48.4 & 65.4 & 64.9 \\
CBERL \cite{meng2024deep} & 79.8 & 82.0 & 56.4 & 57.9 & 26.3 & \textbf{22.2} & 44.2 & 41.4 & 67.1 & 65.7 & 10.2 & 14.7 & 52.8 & 55.3 & 64.4 & 65.5 \\
AdaIGN \cite{tu2024adaptive} & 76.1 & 77.8 & \textbf{59.9} & 60.5 & 22.4 & 15.2 & 44.8 & 43.7 & 66.2 & 64.5 & 33.2 & \textbf{29.3} & 54.3 & 56.2 & 65.3 & 65.8 \\
DER-GCN \cite{ai2024gcn} & 76.8 & 80.6 & 50.5 & 51.0 & 14.8 & 10.4 & 56.7 & 41.5 & 69.3 & 64.3 & 17.2 & 10.3 & 52.5 & \textbf{57.4} & 65.2 & 65.5 \\
FrameERC \cite{li2025frameerc} & 80.7 & 80.0 & 56.4 & 58.1  &  12.3 & 10.5 &  41.2 & 42.5 & 60.8 &  63.4 &  18.4 & 20.1 & 55.1  &  56.3 & 65.5 & 66.1  \\
HiMul-LGG \cite{fu2025himul}       &  77.2 & 75.5 & 55.1 &        53.2 & 10.0 & 9.1 & \textbf{64.4} & \textbf{65.9} & 69.3 & \textbf{68.1} & 8.7 & 9.3 & 52.2 & 52.9 & 66.2  & 65.2  \\
\midrule
DF-GCN & \textbf{81.5} & 80.0 & 56.8 & \textbf{62.7} & \textbf{27.6} & 20.3 & 40.8 & {43.5} & 64.7 & 64.9 & \textbf{34.1} & 26.8 & \textbf{56.1} & 55.4 & \textbf{67.4} & \textbf{67.6} \\
\bottomrule[0.7mm]
\end{tabular}}
\end{table*}

\subsection{Experimental Setup}

All experiments were conducted using Python~3.8 and PyTorch~1.8, and all computations were performed on a single NVIDIA RTX~4090 GPU with 24~GB memory. The proposed model was trained using the AdamW optimizer with an initial learning rate of $5\times10^{-4}$, a weight decay of $1\times10^{-4}$, and the cross-entropy loss function. The batch size was set to 32, and the maximum number of training epochs was set to 50, with early stopping applied based on validation performance. For the model architecture, the hidden dimension of node representations was set to 256, and three stacked graph convolution blocks were used for both SGCODE and DGCODE. The context window size for graph construction was fixed to $w=10$. For the ODE-based modules, the initial state was defined by the encoder output, and the ODE system was solved using a fourth-order Runge--Kutta method with a fixed step size of 0.1. The prompt generation network consisted of a two-layer multilayer perceptron with ReLU activation, and the dimension of the GIV was set to 256. For visualization, t-SNE was applied to the learned utterance-level representations extracted from the test set only. We used the scikit-learn implementation of t-SNE with two output dimensions, random initialization, a perplexity value of 30, and 1000 optimization iterations, while other parameters were set to their default values. To determine the optimal hyperparameters, leave-one-out cross-validation was employed on the validation set. All reported results represent the average performance over 10 independent runs with different random weight initializations. To assess statistical significance, paired $t$-tests were conducted across the 10 runs, and all improvements were found to be statistically significant ($p<0.05$), indicating that the observed performance gains are unlikely to be caused by random variation.


\subsection{Overall Result}
{The experimental results, as shown in Tables \ref{tab:iemocap} and \ref{tab:meld}, clearly demonstrate that the proposed DFGCN method delivers significant performance improvements across various emotion recognition tasks.} This consistent improvement can be primarily attributed to the design of the ODE-injected graph convolutional network, which explicitly models the dynamic evolution of emotional dependencies within discourse interaction networks. By leveraging the global discourse-level information vector as guidance, our model is capable of dynamically fusing multimodal features in a context-aware manner, thereby capturing subtle variations in emotional transitions that traditional static fusion strategies often fail to recognize.

To further illustrate the advantages of our framework, we report the weighted-F1 (W-F1) scores for individual emotion categories. On the IEMOCAP dataset, DFGCN consistently outperforms strong baseline models in critical categories such as happy, neutral, and depressed. These categories are known to be challenging due to their overlapping linguistic and acoustic patterns, and the observed improvements suggest that our model is particularly effective at disentangling such fine-grained distinctions. Similarly, on the MELD dataset, DFGCN achieves the highest W-F1 score for the sad category, while also delivering competitive results in surprise and joy, which are emotions typically sensitive to both verbal and non-verbal cues. These results indicate that the model not only excels in recognizing primary emotions but also generalizes well to subtle and context-dependent emotional states.

While DF-GCN achieves state-of-the-art overall performance on the MELD dataset, particularly in terms of weighted average F1, a closer examination of the per-class results in Table 2 shows that recognition accuracy for certain minority emotion categories, especially Sadness and Joy, remains relatively limited. This limitation does not arise from deficiencies in our dynamic fusion mechanism. Instead, it is primarily driven by two interconnected challenges inherent to the MELD dataset. The first challenge is the severe class imbalance. The Neutral category constitutes more than half of all utterances, whereas Sadness and Joy contribute only about 3 percent and 4.7 percent of the test set. This extreme imbalance restricts the model’s exposure to representative samples of minority emotions and significantly constrains its ability to learn stable and discriminative features, even when the fusion mechanism adapts to contextual cues. The second challenge relates to the highly heterogeneous and context-sensitive nature of multimodal emotion expression in natural dialogue. Sadness may manifest through muted prosodic patterns, subtle or minimal facial expressions, or may be concealed by irony or social politeness. Joy can appear as intense laughter or quiet and restrained expressions of positive affect. These diverse realizations often produce inconsistent or weak alignment among textual, acoustic, and visual modalities, thereby increasing intra-class variability and making minority emotions particularly prone to misclassification as the dominant Neutral class. Although the DGCODE module uses global contextual prompts to adaptively reweight modality contributions and our confusion matrices confirm reduced misclassification of Sadness as Neutral compared with static-fusion baselines, the benefit remains bounded by the scarcity and ambiguity of minority samples. These findings highlight a broader issue common to current multimodal emotion recognition in conversation. Without incorporating dedicated techniques for handling long-tailed distributions, including focal loss, balanced resampling, or decoupled representation learning, even advanced dynamic fusion architectures face inherent challenges in achieving uniformly strong performance across all emotion categories. We recognize this as an important direction for future research.

{Furthermore, it is noteworthy that the performance gains achieved by the model come with only a marginal increase in computational cost when compared to existing graph-based methods.} Although the proposed framework introduces ODE-based propagation and dynamic parameter generation, the overall model complexity and inference time remain comparable to strong graph-based baselines. The integration of ODE dynamics enables efficient representation learning while keeping the computational overhead manageable. This favorable balance between accuracy and efficiency highlights the practicality of DFGCN for real-world multimodal emotion recognition systems, where both predictive performance and computational cost are critical considerations.

\subsection{The Effectiveness of Multimodal Features}

\begin{figure}
	\centering
	\includegraphics[width=1.0\linewidth]{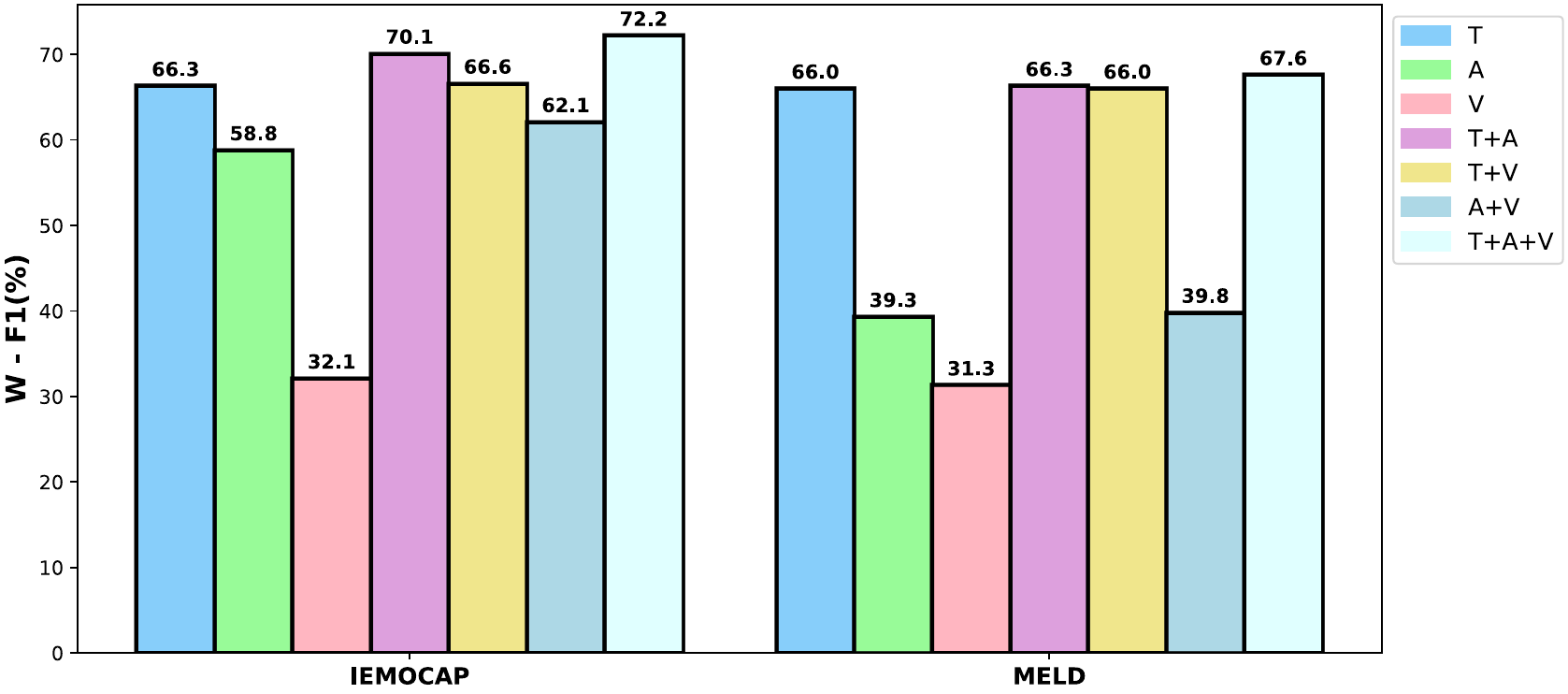}
	\caption{Verifying the effectiveness of multimodal features.}
	\label{fig:figure3}
\end{figure}

{To assess the impact of different combinations of modality features on emotion recognition performance, we performed a detailed ablation study, analyzing the contributions of text, audio, and video features both individually and in various combinations.} Specifically, we designed experiments in which the model was provided with different subsets of modalities, and we carefully examined the resulting changes in recognition accuracy. The outcomes of these experiments are summarized in Fig. \ref{fig:figure3}.

In the first set of experiments, where each modality was tested in isolation, we observed that the text modality consistently delivered the highest recognition accuracy compared with the audio and video modalities. This finding indicates that textual features contain the most discriminative cues for emotion classification, likely because they directly encode semantic content and contextual information that are critical for distinguishing between nuanced emotional states. By contrast, audio and visual features, although useful, proved less effective when used independently, as they tend to capture paralinguistic or nonverbal signals that may be more ambiguous or context-dependent.

Subsequently, we evaluated the performance of the model when combining two modalities. The results revealed that both the text–audio combination and the text–video combination yielded significant improvements over the single-modality settings. These enhancements suggest that the integration of multiple modalities enables the model to capture complementary information. While the text modality provides a strong semantic foundation, the audio modality contributes prosodic and acoustic cues, and the video modality adds visual expressions such as facial movements and gestures. The fusion of these heterogeneous features therefore strengthens the model’s ability to accurately infer emotions, particularly in cases where a single modality may provide incomplete or ambiguous signals.

Finally, when all three modalities (e.g., text, audio, and video) were used jointly, the model achieved its best overall performance. This outcome highlights the fundamental advantage of multimodal learning in emotion recognition tasks. By simultaneously exploiting information from linguistic, acoustic, and visual channels, the model is able to construct a more comprehensive and context-aware representation of emotional states, leading to predictions that are both more accurate and more robust. 

\subsection{Sensitivity Analysis of Graph Construction Hyperparameters}

Fig.~\ref{fig:figure5} presents the sensitivity analysis of DF-GCN with respect to two key graph construction hyperparameters: the similarity threshold $\theta$ and the context window size $w$. Results {from} both the IEMOCAP and MELD datasets indicate that the proposed model {maintains stable performance over a wide range of parameter values, suggesting that DF-GCN is not overly sensitive to the selection of these hyperparameters.} As shown in Fig.~1(a), varying the similarity threshold $\theta$ results in only moderate performance fluctuations. When $\theta$ is too small, the constructed graph may become overly dense and introduce noisy connections, whereas excessively large values of $\theta$ tend to produce sparse graphs that limit effective information propagation. The model achieves consistently strong performance when $\theta$ is set to an intermediate range, indicating a balanced trade-off between graph connectivity and noise suppression. Fig.~1(b) illustrates the effect of the context window size $w$. A small window size restricts the available conversational context, while an overly large window may introduce redundant or less relevant interactions. The results show that DF-GCN maintains robust performance across different window sizes, with moderate values of $w$ providing an effective balance between contextual coverage and modeling efficiency.

\begin{figure}[htbp]
	\centering
	\subfloat[Similarity threshold]{\includegraphics[width=0.45\linewidth]{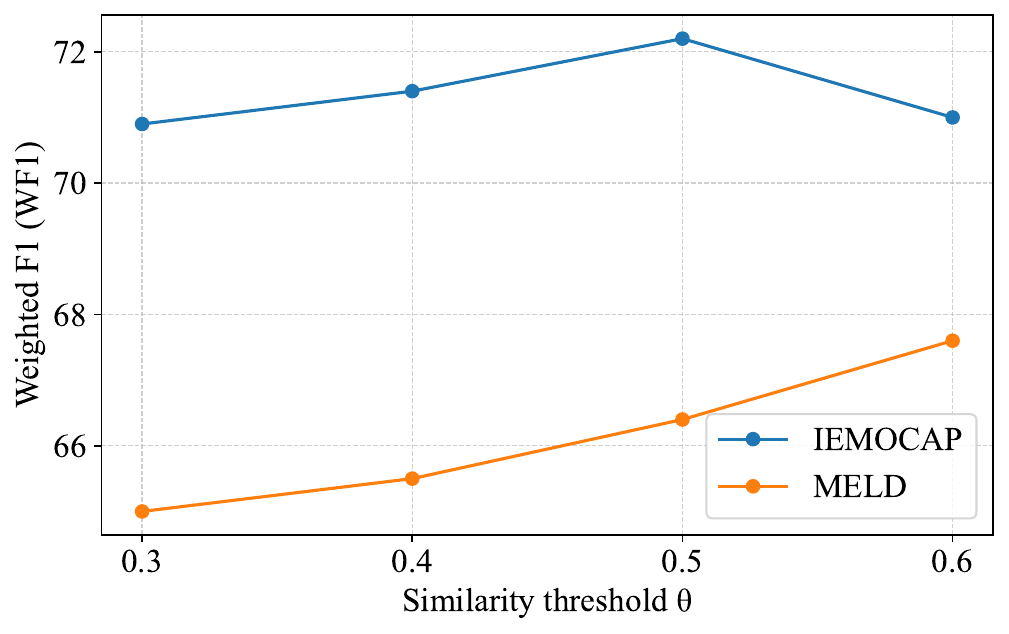}}%
		\label{fig:embed_visual_emo_initial_iemocap6}
	\hfil
	\subfloat[Context window]{\includegraphics[width=0.45\linewidth]{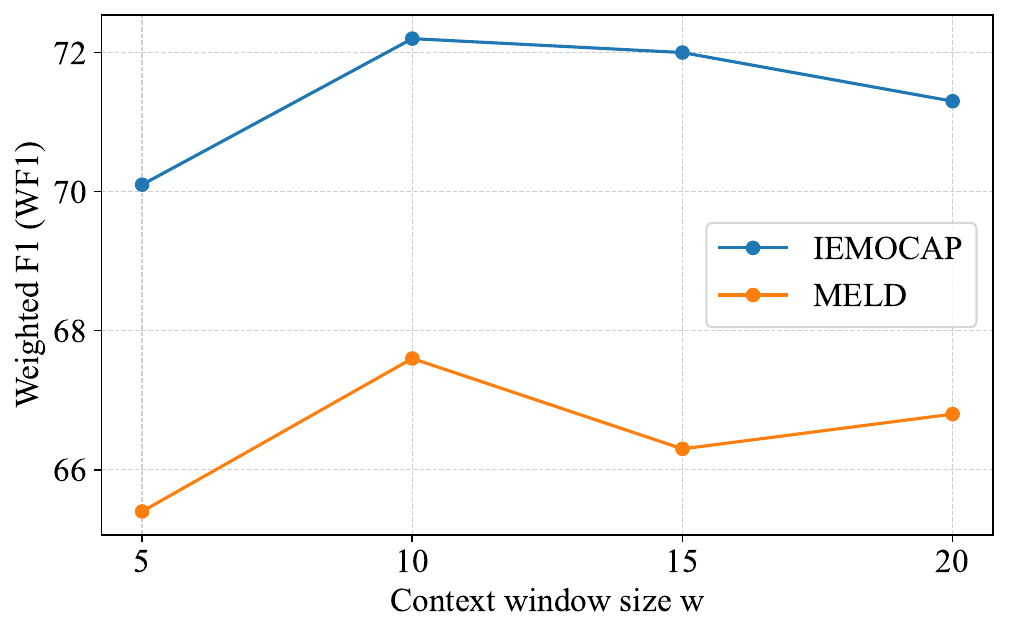}%
		\label{fig:embed_visual_emo_mmgcn_iemocap6}}
	\vfil
	\caption{Sensitivity analysis of DF-GCN with respect to key graph construction hyperparameters. (a) Effect of the similarity threshold $\theta$ on WF1. (b) Effect of the context window size $w$ on WF1.}
	\label{fig:figure5}
\end{figure}

\subsection{Error Analysis}

\begin{figure*}[htbp]
    \centering
    \begin{minipage}[t]{0.24\linewidth}  
        \centering
        \includegraphics[width=1\linewidth]{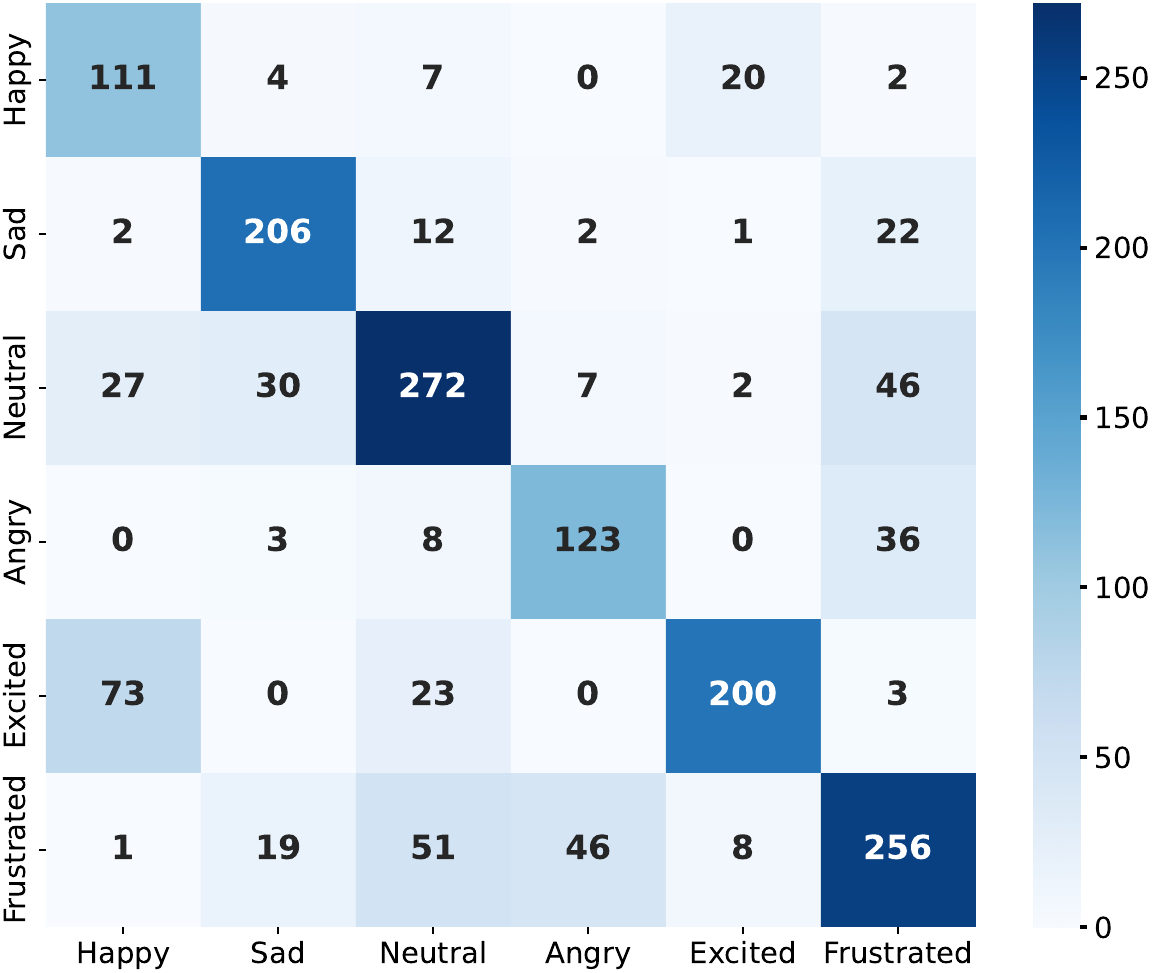}
        \label{fig:confusion_matrix_iemocap}
        \parbox{\linewidth}{\centering (a) DF-GCN (IEMOCAP)}
    \end{minipage}
    \hspace{0\linewidth}  
    \begin{minipage}[t]{0.24\linewidth}  
        \centering
        \includegraphics[width=1\linewidth]{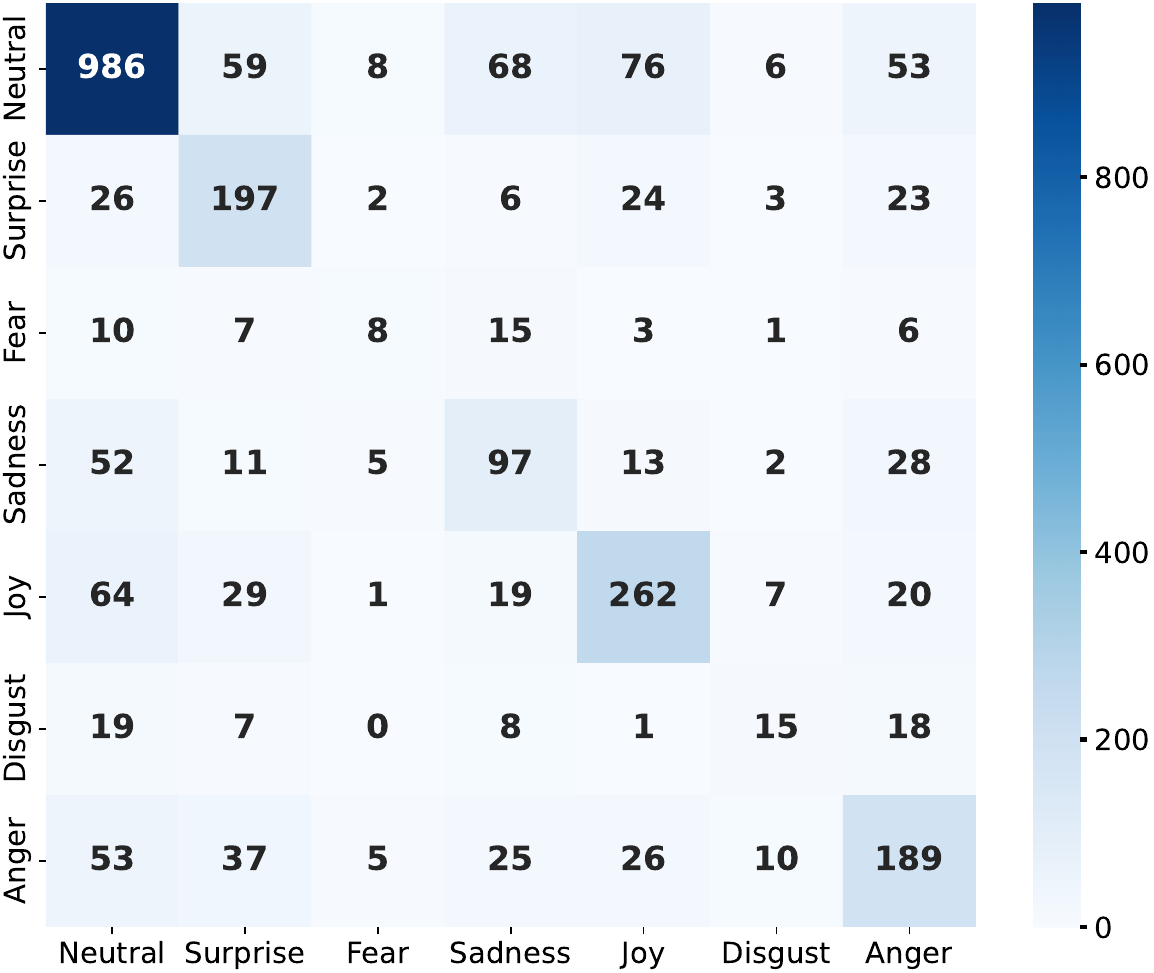}
        \label{fig:confusion_matrix_meld}
        \parbox{\linewidth}{\centering (b) DF-GCN (MELD)}
    \end{minipage}
        \begin{minipage}[t]{0.24\linewidth}  
        \centering
        \includegraphics[width=1\linewidth]{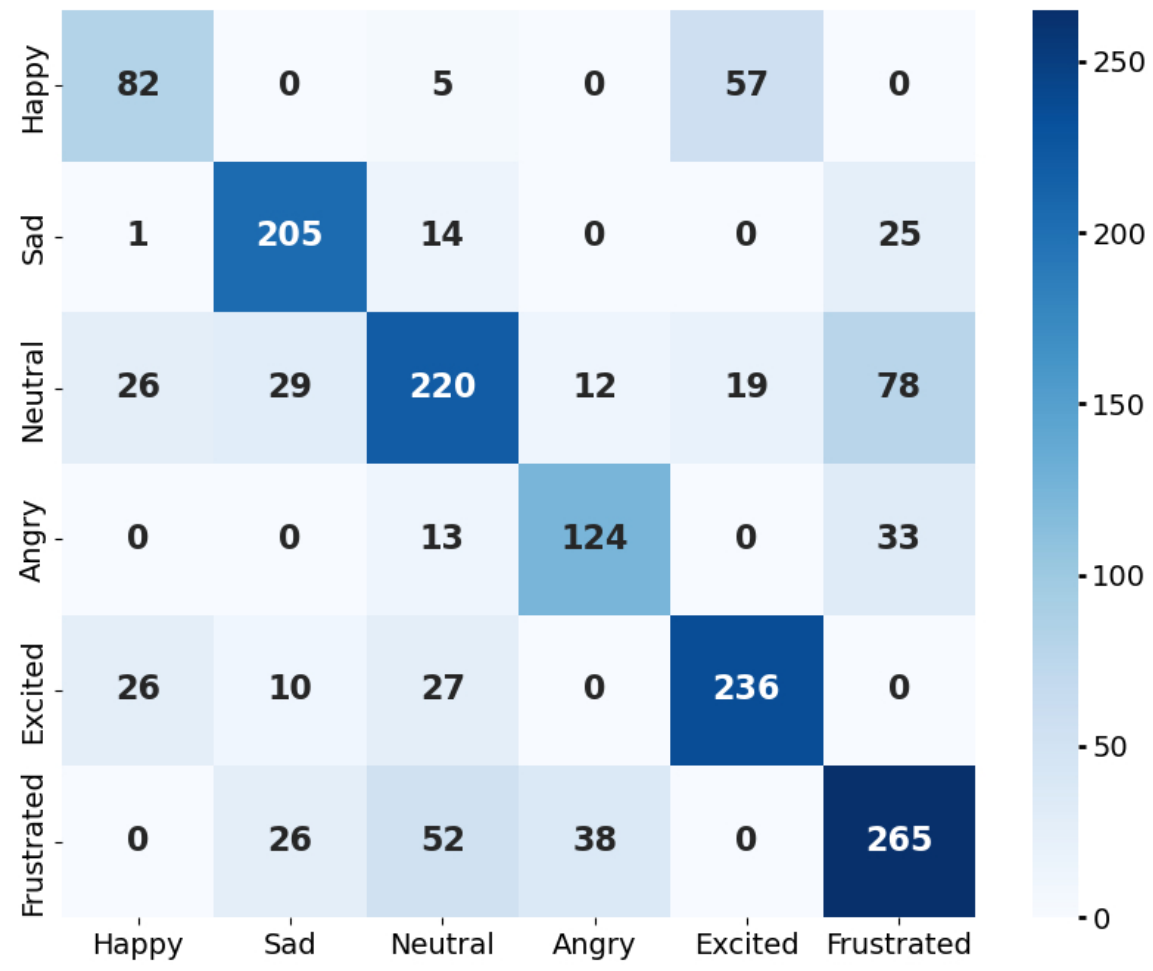}
        \label{fig:confusion_matrix_meld}
        \parbox{\linewidth}{\centering (c) DER-GCN (IEMOCAP)}
    \end{minipage}
        \begin{minipage}[t]{0.24\linewidth}  
        \centering
        \includegraphics[width=1\linewidth]{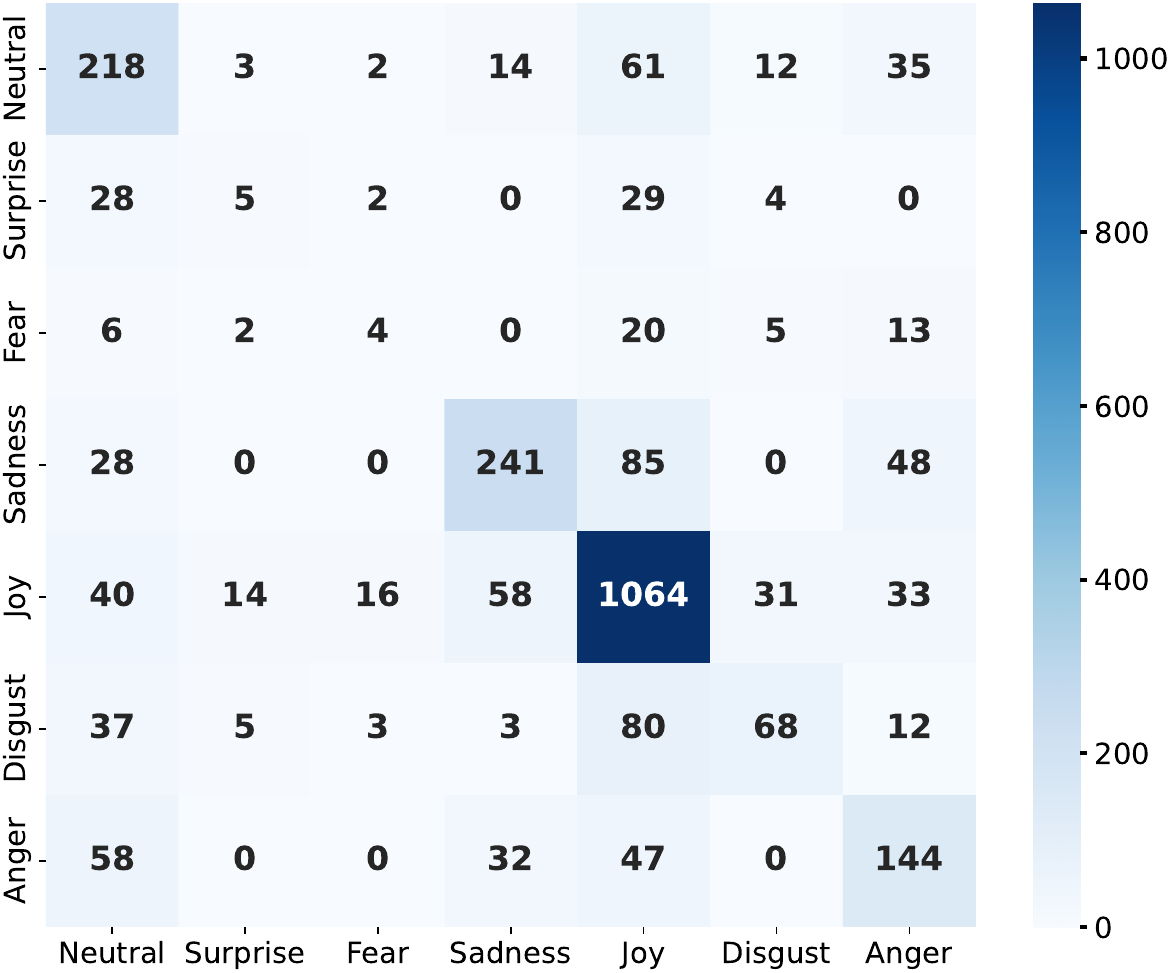}
        \label{fig:confusion_matrix_meld}
        \parbox{\linewidth}{\centering ({d}) DER-GCN (MELD)}
    \end{minipage}
    \caption{Confusion matrices of emotion classification results on the IEMOCAP and MELD test sets. Subfigures (a) and (b) present the performance of the proposed DF-GCN model on IEMOCAP and MELD, respectively, while subfigures (c) and (d) show the results of the baseline DER-GCN under the same settings. Darker diagonal blocks indicate higher accuracy in correctly identifying specific emotion categories, whereas off-diagonal values represent misclassifications.}
    \label{fig:figure4}
\end{figure*}

{To better understand} the model’s limitations, {we analyzed the confusion matrices of} the IEMOCAP and MELD test sets. As shown in Figs. \ref{fig:figure4} (a) and (b), DFGCN demonstrates stronger diagonal concentration compared with DER-GCN, yet it still exhibits several notable misclassification patterns. On the IEMOCAP dataset, DFGCN often predicts excited as happy and confuses angry with frustrated or neutral. While similar errors are also present in DER-GCN (Fig. \ref{fig:figure4} (c)), the overall distribution of DFGCN shows fewer cross-category confusions, particularly for the sad and frustrated classes, indicating its improved ability to capture fine-grained distinctions.

On the MELD dataset, DFGCN also reduces confusion compared to DER-GCN, especially in the recognition of joy and sadness. Nevertheless, challenges remain due to class imbalance: the dominance of the neutral category biases predictions, leading to frequent misclassification of minority emotions, such as fear into neutral. Moreover, categories like fear and disgust, which are underrepresented in MELD, remain difficult to classify correctly for both models. Although DFGCN leverages dynamic weight allocation to mitigate imbalance, its recognition of these minority classes is still limited, albeit consistently superior to DER-GCN.

\subsection{Computational Efficiency Analysis}

Table \ref{tab:efficiency} presents a comparison of model size and inference efficiency among different graph-based approaches on the IEMOCAP and MELD datasets. Overall, the proposed DF-GCN achieves a favorable balance between recognition performance and computational efficiency. From the perspective of model complexity, DF-GCN maintains a moderate parameter scale compared with existing methods. While it introduces additional components such as ODE-based graph propagation and dynamic parameter generation, the overall parameter count remains comparable to mainstream graph-based models and is substantially smaller than that of more complex relational graph architectures. This indicates that the proposed design avoids excessive parameter growth while enhancing modeling capacity. In terms of inference efficiency, DF-GCN demonstrates competitive runtime performance across both datasets. Its inference speed is on par with commonly used graph convolutional models and notably faster than heavier graph-based baselines that rely on complex relational modeling. This suggests that the additional computations introduced by dynamic parameter generation and ODE integration are effectively controlled and do not lead to a prohibitive increase in inference time. Overall, these results show that the performance gains achieved by DF-GCN come with only marginal additional computational cost. The lightweight prompt generation network and the fixed-depth ODE integration ensure that the proposed model remains efficient in practice.

\begin{table}[htbp]
\centering
\caption{Comparison of model parameters and inference time on IEMOCAP and MELD datasets.}
\label{tab:efficiency}
\setlength{\tabcolsep}{0.5mm}{
\begin{tabular}{lccc}
\hline
\multirow{2}{*}{Model} & \multirow{2}{*}{\# Params.} & IEMOCAP            & MELD               \\ \cline{3-4} 
                       &                               & Inference Time & Inference Time \\ \hline
MMGCN                  & 5.63M                         & 53.7s              & 75.3s              \\
RGAT                   & 15.28M                        & 68.5s              & 146.3s             \\
DER-GCN                & 78.59M                        & 125.5s             & 189.7s             \\
M3Net                  & 7.88M                         & 57.0s             & 87.6s             \\
DF-GCN (Ours)          & 9.19M                         & 60.2s             & 94.1s             \\ \hline
\end{tabular}}
\end{table}

{\subsection{Stability and Robustness Analysis}}

{As illustrated in Table \ref{tab:robu}, our proposed DF-GCN not only achieves the state-of-the-art (SOTA) performance in terms of average Accuracy and F1-score across both IEMOCAP and MELD datasets but also exhibits superior stability compared to all baseline methods.Specifically, on the IEMOCAP dataset, DF-GCN yields the lowest standard deviation for both metrics, significantly outperforming other competitive models like M3Net and AdaIGN. A similar trend is observed on the MELD dataset, where DF-GCN maintains an remarkably low variance, while the variances of most baselines range from $0.45$ to $1.58$.The consistently smaller standard deviation after 10 independent runs indicates that DF-GCN is less sensitive to random initialization and noise within the multimodal data. These results strongly demonstrate the robustness of our architecture, confirming that the performance gains of DF-GCN are reliable and reproducible rather than the result of advantageous random seeding.}

\begin{table}[htbp]
\centering
\caption{{Average performance after 10 runs on the IEMOCAP and MELD datasets. The values reported in each cell represent the average Acc/F1 $\pm$ std. The best result
in each column is in bold.}}
\label{tab:robu}
\setlength{\tabcolsep}{0.5mm}{
\begin{tabular}{lcccc}
\hline
\multirow{2}{*}{Methods} & \multicolumn{2}{c}{IEMOCAP} & \multicolumn{2}{c}{MELD} \\ \cline{2-5} 
                         & Acc             & F1        & Acc         & F1         \\ \hline
QMNN                     & 60.8 $\pm$ 0.87     & 59.9 $\pm$ 1.23    & 60.8  $\pm$ 0.45     & 58.0 $\pm$ 1.58     \\
CoMPM                    & 64.2 $\pm$  0.19         & 67.3 $\pm$  1.02   & 64.3 $\pm$ 0.66      & 63.0  $\pm$ 1.41    \\
EmoBERTa                 & 67.3 $\pm$  0.33         & 67.3 $\pm$  1.15   & 64.1 $\pm$  0.92     & 63.3  $\pm$  0.54   \\
MMGCN                    & 66.8 $\pm$ 1.37          & 66.8 $\pm$  0.28   & 58.3  $\pm$  1.49    & 58.4  $\pm$  0.71   \\
M3Net                    & 72.5 $\pm$ 1.08          & 71.1 $\pm$ 1.55    & 63.9  $\pm$   0.83   & 65.8  $\pm$  1.29   \\
MGLRA                    & 71.3 $\pm$  0.41         & 70.1 $\pm$ 0.96    & 65.4 $\pm$  1.12     & 64.9  $\pm$ 0.62    \\
CBERL                    & 69.4 $\pm$ 1.44          & 69.3 $\pm$ 1.33    & 64.4 $\pm$ 0.78      & 65.5  $\pm$  1.05   \\
AdaIGN                   & 72.1 $\pm$  0.51         & 70.7 $\pm$ 1.52    & 65.3 $\pm$  1.19     & 65.8  $\pm$  0.89   \\
DER-GCN                  & 69.7 $\pm$  1.46         & 69.4 $\pm$  1.31   & 65.2  $\pm$  0.68    & 65.5  $\pm$  1.01 \\
DF-GCN                   & \textbf{73.4} $\pm$  0.21         & \textbf{72.2} $\pm$  0.17   & \textbf{67.4}  $\pm$  0.13    & \textbf{67.6}  $\pm$  0.25
\\ \hline
\end{tabular}}
\end{table}

\subsection{Ablation Experiment}

To analyze the contribution of each component in DF-GCN, we conducted a series of ablation experiments on the IEMOCAP and MELD datasets. As shown in Table \ref{tab:table2}, DF-GCN consistently outperforms all ablated variants in terms of both Accuracy (Acc) and Weighted F1 score (W-F1), demonstrating the effectiveness of the complete model architecture. These results indicate that removing any individual component leads to a noticeable performance drop, confirming that each module contributes meaningfully to the overall performance. In particular, removing the GIV results in a clear degradation in performance, highlighting its critical role in capturing and integrating global contextual information across the conversation. GIV enables the model to maintain a holistic understanding of emotional context, which is essential for robust multimodal emotion recognition. Moreover, eliminating the PGN or DGCODE also leads to reduced performance, underscoring the importance of dynamically adapting model parameters to emotion-specific contexts. The joint operation of PGN and DGCODE allows the model to generate adaptive weights conditioned on emotional dynamics, thereby enhancing classification accuracy and robustness. 

\begin{table}[H]
\caption{Ablation results of DF-GCN on the IEMOCAP and MELD benchmark dataset.}
\label{tab:table2}
\setlength{\tabcolsep}{2.7mm}{
\begin{tabular}{ccccccc}
\toprule[0.7mm]
\multirow{2}{*}{GIV} & \multirow{2}{*}{PGN} & \multirow{2}{*}{DGCODE} & \multicolumn{2}{l}{IEMOCAP} & \multicolumn{2}{l}{MELD} \\ \cline{4-7} 
                     &                      &                         & Acc.          & F1          & Acc.         & F1        \\ \midrule[0.4mm]
+ & + & + & \textbf{73.4} & \textbf{72.2} & \textbf{67.4} & \textbf{67.6} \\
- & + & + & 71.3 & 71.6 & 67.2 & 65.8 \\
+ & - & +  & 70.6 & 70.8 & 66.9 & 64.8 \\
+ & + & - & 69.4  & 68.7 & 65.4 & 64.4 \\
- & - & +  & 70.6 & 70.8 & 66.8 & 65.0 \\
- & + & - & 68.9 & 68.4 & 64.1 & 63.5  \\
+ & - & - & 68.5 & 68.0 & 63.3 & 62.4 \\
- & - & - & 66.7 &  67.0 & 62.3 & 61.6      \\ \bottomrule[0.7mm]
\end{tabular}}
\end{table}

\begin{figure*}[htbp]
	\centering
	\subfloat[Initial (IEMOCAP)]{\includegraphics[width=0.24\linewidth]{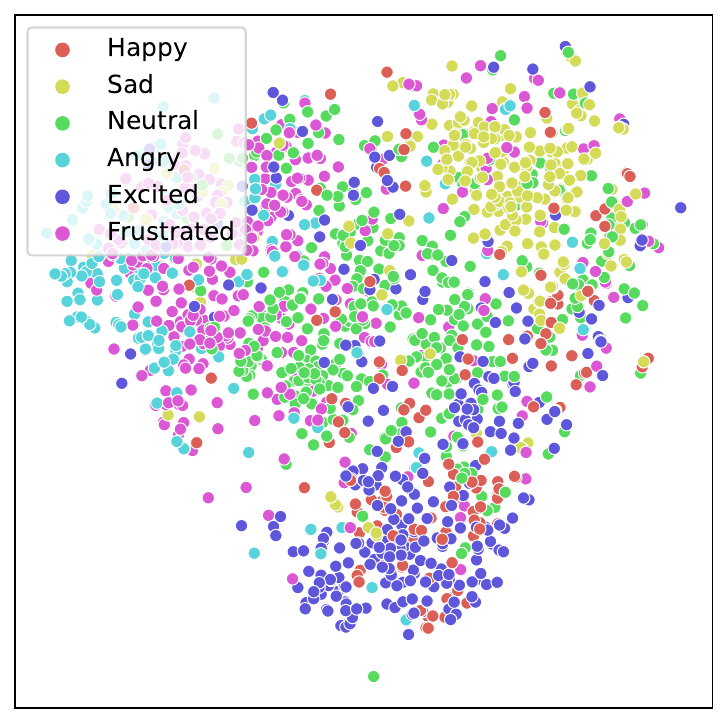}%
		\label{fig:embed_visual_emo_initial_iemocap6}}
	\hfil
	\subfloat[MMGCN (IEMOCAP)]{\includegraphics[width=0.24\linewidth]{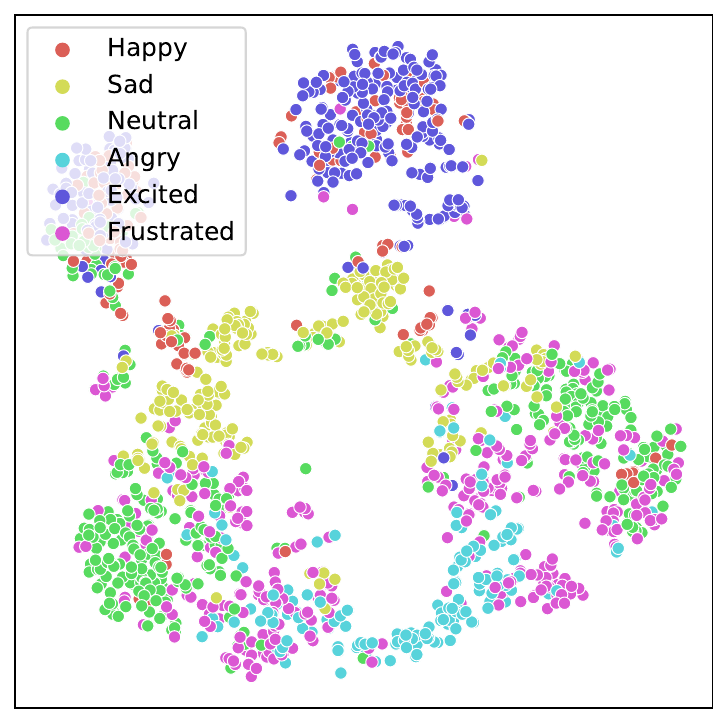}%
		\label{fig:embed_visual_emo_mmgcn_iemocap6}}
	\hfil
	\subfloat[M3Net (IEMOCAP)]{\includegraphics[width=0.24\linewidth]{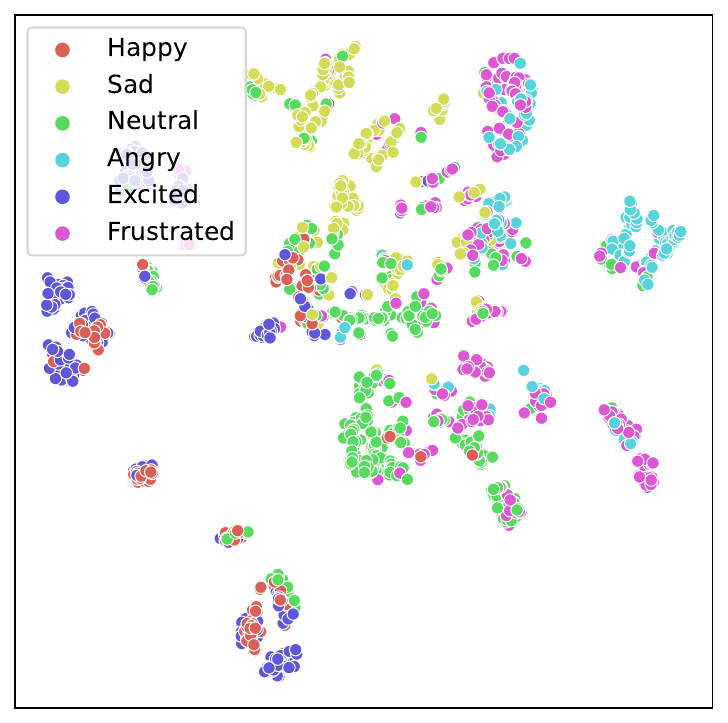}%
		\label{fig:embed_visual_emo_m3net_iemocap6}}
	\hfil
	\subfloat[DF-GCN (IEMOCAP)]{\includegraphics[width=0.24\linewidth]{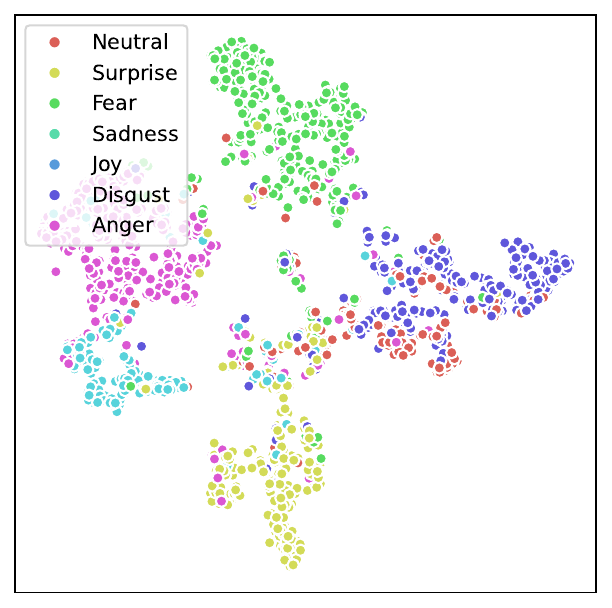}%
		\label{fig:embed_visual_emo_graphsmile_iemocap6}}
	\vfil
	\subfloat[Initial (MELD)]{\includegraphics[width=0.24\linewidth]{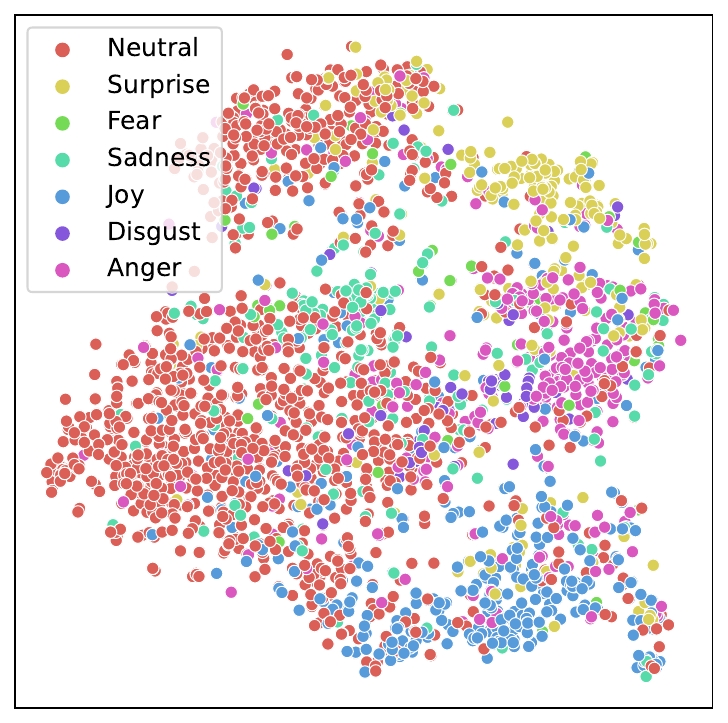}%
		\label{fig:embed_visual_emo_initial_meld}}
	\hfil
	\subfloat[MMGCN (MELD)]{\includegraphics[width=0.24\linewidth]{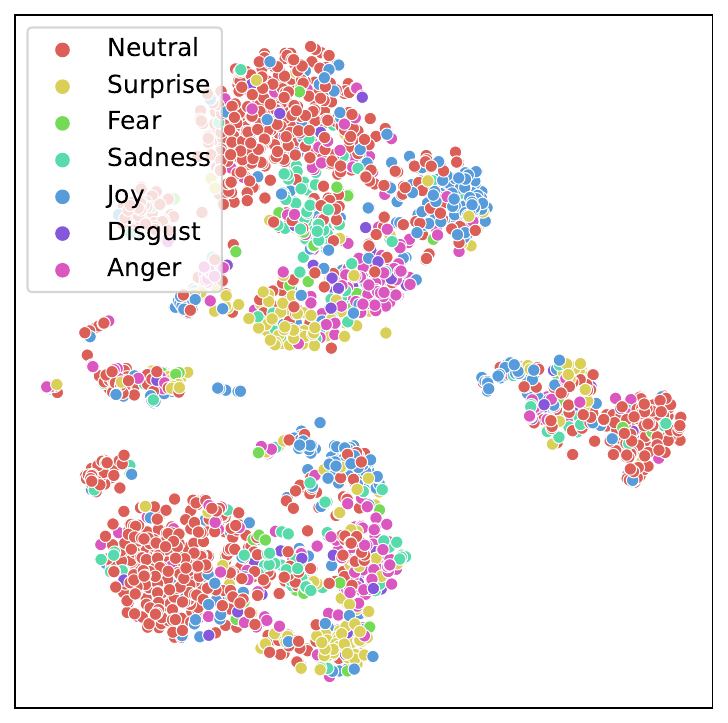}%
		\label{fig:embed_visual_emo_mmgcn_meld}}
	\hfil
	\subfloat[M3Net (MELD)]{\includegraphics[width=0.24\linewidth]{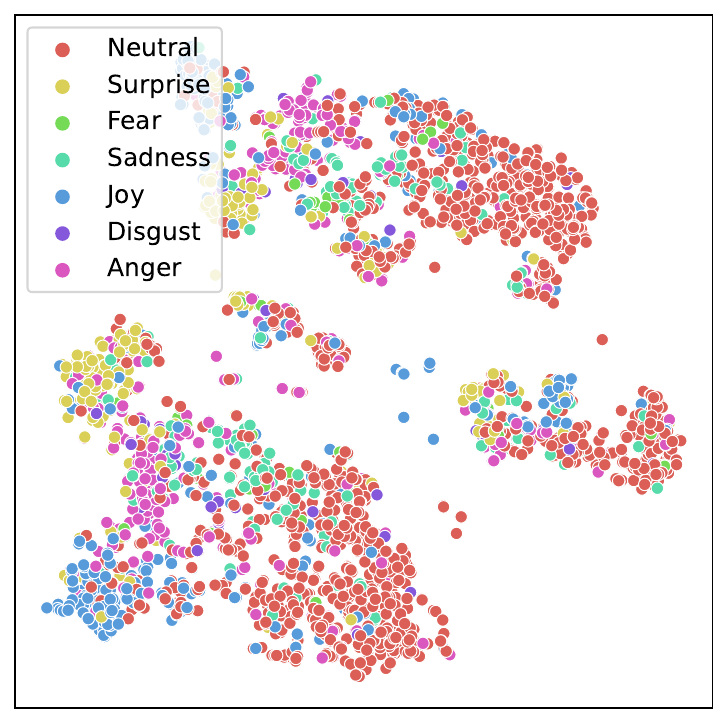}%
		\label{fig:embed_visual_emo_m3net_meld}}
	\hfil
	\subfloat[DF-GCN (MELD)]{\includegraphics[width=0.24\linewidth]{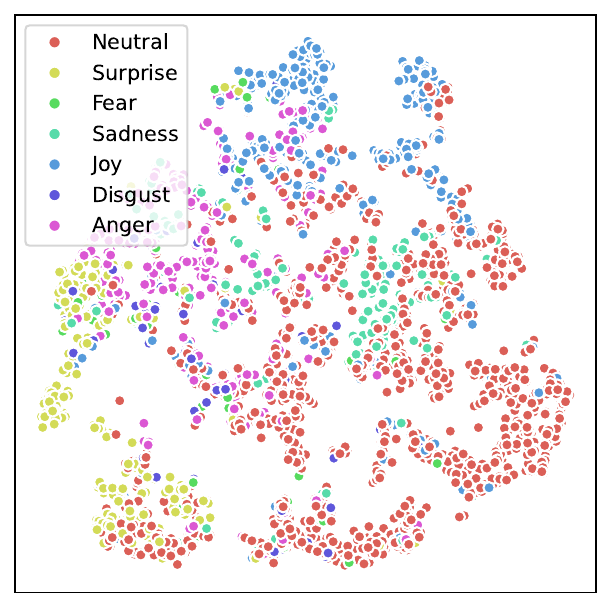}%
		\label{fig:embed_visual_emo_graphsmile_meld}}
	\vfil
	\caption{Visualization of the learned embeddings.}
	\label{fig:figure5}
\end{figure*}

\subsection{Visualization}
To more intuitively demonstrate the classification effect of the DF-GCN method on the MERC task, we used the T-SNE technology to reduce the dimension of the generated sentence vectors and visualize them. As shown in Fig. \ref{fig:figure5}, on the IEMOCAP dataset, the DF-GCN model shows excellent classification performance. We can see that different emotion categories form obvious cluster distributions, and the distances between different clusters are large, indicating that DF-GCN can well capture the differences between different emotions. In contrast, although the MMGCN model can also distinguish samples of different emotion categories to some extent, its classification performance is obviously inferior to that of DF-GCN. From the visualization results, it can be seen that the distribution of samples generated by MMGCN is relatively chaotic, the boundaries between different emotion categories are not clear enough, and samples of some categories even overlap. For example, samples of happy and neutral emotions are distributed close to each other in the figure, which may cause confusion during classification. Furthermore, we also compared the classification effect of the M3Net model on the IEMOCAP dataset. Similar to DF-GCN, M3Net also shows good classification performance and can clearly separate samples of different emotion categories. It can be seen that the samples generated by M3Net are more concentrated, the boundaries between categories are clear, and the clustering effect is comparable to that of DF-GCN. To further verify the generalization ability of DF-GCN, we conducted similar experiments on the MELD dataset and observed the same phenomenon as that on the IEMOCAP dataset. DF-GCN also showed excellent classification performance, and samples of different emotion categories were effectively separated in the two-dimensional space.

\section{Conclusion}
 {This paper presents DF-GCN, a robust framework for recognizing multimodal emotional features in conversations.} Specifically, DF-GCN injects ordinary differential equations into GCN to explore the dynamics of emotion dependencies in utterance interaction networks and  leverages the prompts generated by the global information vector of the utterance to guide the dynamic fusion of multimodal features. This allows our model to dynamically change parameters when processing each utterance feature, so that different network parameters can be equipped for different emotion categories in the inference stage, thereby achieving more flexible emotion classification and enhancing the generalization ability of the model. Extensive experiments on two public multimodal conversation datasets demonstrate that the proposed DF-GCN model achieves superior performance and explicitly reveals the uncertainty of emotion variations through dynamics fusion. To the best of our knowledge, we make the first attempt to assign different weights to different emotion categories in the inference stage to achieve more effective information fusion.

 \section*{CRediT authorship contribution statement}

\textbf{Tao Meng}: Conceptualization, Methodology, Investigation, Data curation, Writing - Original Draft. \textbf{Weilun Tang}: Supervision, Investigation \& Review. \textbf{Yuntao Shou}: Supervision, Investigation, Writing - Review \& Editing. \textbf{Yilong Tan}: Writing - Review \& Editing. \textbf{Jun Zhou}: Supervision, Investigation, Writing - Review \& Editing. \textbf{Wei Ai}: Supervision, Investigation, Writing- Review \& Editing. \textbf{Keqin Li}: Supervision, Investigation, Writing - Review \& Editing.

\section*{Declaration of Competing Interest}

The authors declare that they have no known competing financial interests or personal relationships that could have appeared to influence the work reported in this paper.

\section*{Data availability}

Data will be made available on request.

\section*{Acknowledgements}

The authors deepest gratitude goes to the anonymous reviewers and AE for their careful work and thoughtful suggestions that have helped improve this paper substantially. This
work is supported by National Natural Science Foundation
of China (Grant No. 69189338), Excellent Young Scholars
of Hunan Province of China (Grant No. 22B0275), and program of Research on Local Community Structure Detection Algorithms in Complex Networks (Grant No. 2020YJ009).

\bibliographystyle{ieeetr}
\bibliography{refs}

\end{document}